\pdfoutput=1
\documentclass[11pt]{article}

\usepackage[]{acl}

\usepackage{times}
\usepackage{latexsym}
\usepackage{colortbl}
\usepackage{bbding}
\usepackage{pdfpages}
\usepackage[T1]{fontenc}
\usepackage[utf8]{inputenc}

\usepackage{microtype}

\usepackage{inconsolata}

\usepackage{graphicx}

\usepackage{booktabs}
\usepackage[intlimits]{amsmath}  % place the subscripts and superscripts in the right position
\usepackage[colorinlistoftodos,size=tiny,disable]{todonotes}
\usepackage{caption}
\usepackage{pifont}
\usepackage{subcaption}
\usepackage{multirow}
\usepackage{enumitem}
\usepackage{makecell}
\usepackage{tabularx}
\usepackage{soul}
\usepackage{amssymb}
\usepackage[rgb,dvipsnames]{xcolor}
\usepackage{afterpage}

\newcommand{\se}[1]{\textcolor{black}{#1}}
\newcommand{\senew}[1]{\textcolor{black}{#1}} % blue
\newcommand{\yc}[1]{\textcolor{black}{#1}} % orange
\newcommand{\ycr}[1]{\textcolor{black}{#1}} % orange
\newcommand{\yct}[1]{\textcolor{black}{#1}}
\newcommand{\yccr}[1]{\textcolor{black}{#1}}
\newcommand{\data}[0]{\textsc{LOGIC}}
\newcommand{\cor}[1]{\textcolor{red}{\textbf{#1}}}
\newcommand{\incor}[1]{\textcolor{OliveGreen}{\textbf{#1}}}

\title{\includegraphics[scale=0.05]{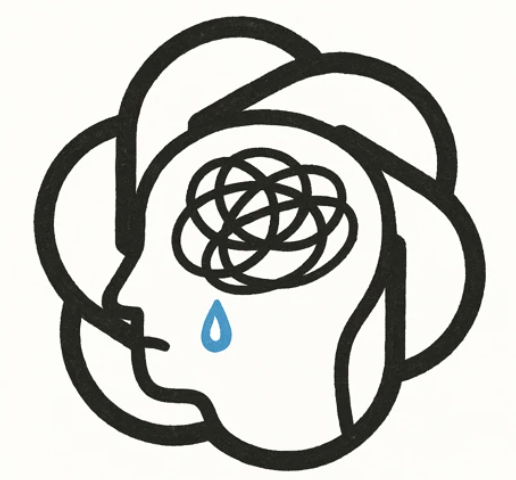} Emotionally Charged, Logically Blurred: \\
AI-driven Emotional Framing \ycr{Impairs Human %Reasoning in 
Fallacy Detection}
}

\author{Yanran Chen$^{1}$, Lynn Greschner$^{2}$, Roman Klinger$^{2}$, Michael Klenk$^{3}$, Steffen Eger$^{1}$ \\
 $^{1}$NLLG, University of Technology Nuremberg (UTN), Germany \\\url{https://nl2g.github.io/}\\
 $^{2}$Fundamentals of Natural Language Processing, University of Bamberg, Germany \\
 $^{3}$Ethics and Philosophy of Technology, Delft University of Technology, Netherlands
  }

\begin{document}

\maketitle

\begin{abstract}
\ycr{Logical fallacies are common in public communication and can mislead audiences; fallacious arguments may still appear convincing despite lacking  
soundness, because convincingness is inherently subjective. We present the first \yct{computational} study of how emotional framing interacts with fallacies and convincingness, using large language models (LLMs) to systematically change emotional appeals in fallacious arguments. We benchmark eight LLMs on injecting emotional appeal into fallacious arguments while preserving their logical structures, then use the best models to generate stimuli for a human study.  
Our results show that LLM-driven emotional framing reduces human fallacy detection in F1 by 14.5\%  
on average. Humans perform better in fallacy detection when perceiving enjoyment than fear or sadness, and these three emotions also correlate with significantly higher convincingness compared to neutral or other emotion states. Our work has implications for AI-driven emotional manipulation in the context of fallacious argumentation.}
\end{abstract}

\section{Introduction}\label{sec:introduction}

The ability to recognize fallacious reasoning is a critical component of robust argumentation, especially in an age where misinformation and 
manipulation proliferate across digital platforms \citep{sahai-etal-2021-breaking,macagno2022argumentation,helwe-etal-2024-mafalda}, and human reasoning can be vulnerable to such fallacies. 
Fallacy detection has therefore become a growing area of interest in natural language processing (NLP) \citep[e.g.,][]{goffredo-etal-2023-argument,lei-huang-2024-boosting,li-etal-2024-reason}%{goffredo-etal-2023-argument,ruiz-dolz-lawrence-2023-detecting,lei-huang-2024-boosting,robbani-etal-2024-flee,li-etal-2024-reason}
. Existing computational approaches, however, primarily target logical structures \yct{of fallacies} and argumentation schemes, 
largely overlooking the emotional dimension of argumentation.\footnote{\yct{We consider emotions as an independent dimension of arguments. }\se{There is some work, especially in the NLP community, that treats emotions themselves as a fallacy \citep[e.g.,][]{li-etal-2024-reason,helwe-etal-2024-mafalda,jin-etal-2022-logical}. We think this is not always sound, as emotions can be justified when there are sufficient reasons for having that emotion \citep{manolescu2006normative,deonna2012justified,echeverri2019emotional}---if the house is burning, it is justified to experience the emotion of fear.}} 
\ycr{Moreover, human performance on those fallacy datasets has rarely been studied, even though it is essential for establishing human baselines when benchmarking AI. \citet{helwe-etal-2024-mafalda} provide the only human evaluation to date, but it is limited to just 20 samples.}

\begin{figure}[!t]
    \centering
    \includegraphics[width=\linewidth]{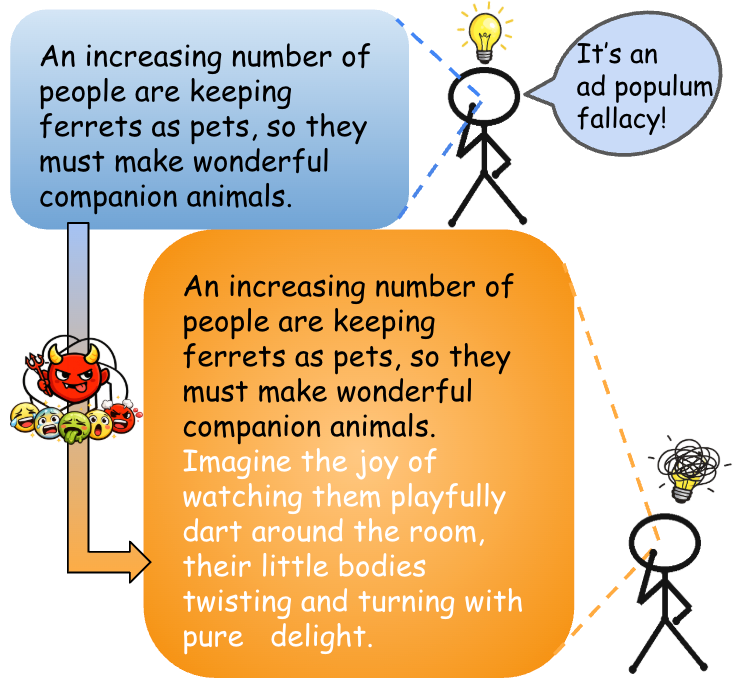}
    \caption{Overview of our approach. We start from fallacious arguments and use LLMs to apply emotional framing strategies to evoke certain emotions while preserving the premise–conclusion structure of the original arguments. We then evaluate human fallacy detection on the original versus LLM emotionally framed arguments. The figure illustrates an example where \emph{enjoyment} is introduced into an \emph{ad populum} argument using the \emph{imagery} strategy; the white text is added by an LLM for emotional framing.}
    \label{fig:overview}
\end{figure}

Findings from psychology and philosophy suggest that emotions play certain roles in reasoning. Emotional content can impact accuracy in deductive reasoning \citep{blanchette2006effect,blanchette2014does}. Emotions can also influence risk perception \citep{lerner2001fear}%{lerner2001fear,slovic2005affect}
, attention \citep{jung2014emotions}, 
%{jung2014emotions,fredrickson2004broaden}
and confidence in judgment \citep{schwarz1983mood}, thereby shaping the cognitive process in logical reasoning and decision-making. Beyond reasoning, emotions are also central to persuasion: emotional appeal has long been recognized in philosophy and argumentation theory \citep{kennedy1991theory,konat2024pathos,benlamine2015emotions}, and several NLP works link emotional aspects to judgments of convincingness \citep[e.g.,][]{wachsmuth-etal-2017-computational,habernal2017argumentation,chen-eger-2025-emotions}%
%{wachsmuth-etal-2017-computational,mirzakhmedova2024large,habernal2017argumentation,ziegenbein-etal-2023-modeling,habernal-gurevych-2016-argument,greschner2024fearfulfalconsangryllamas,chen-eger-2025-emotions}
. Despite these findings, the interaction between emotions, fallacies, and convincingness remains unexplored in computational or human-centered NLP research. 

This gap is particularly concerning in the context of AI-generated text. Large language models (LLMs) are now widely deployed in everyday applications and have demonstrated strong capabilities in synthetic argument generation \citep{schiller-etal-2021-aspect,mouchel2024logicalfallacyinformedframeworkargument,pauli-etal-2025-measuring}.
%{schiller-etal-2021-aspect,li-etal-2024-reason,mouchel2024logicalfallacyinformedframeworkargument,pauli-etal-2025-measuring} 
This raises risks: malicious actors could exploit LLMs to  
create persuasive yet logically unsound content for misinformation, propaganda, or manipulation. 
At the same time, LLMs also provide a unique opportunity to systematically  
\ycr{change} emotional framing while preserving logical content, enabling controlled investigation of its effects on human reasoning \citep{chen-eger-2025-emotions}.

In this paper, we present, to our knowledge, the first computational study that examines the interaction between (AI-driven) emotional framing, fallacy detection, and argument convincingness\yct{, with the main focus on how emotional framing shapes human ability to detect fallacies}. 
\ycr{To establish a controlled setting, we adapt the paradigm of \citet{chen-eger-2025-emotions} to construct fallacious arguments that differ only in emotional tone, and compare human fallacy-detection performance on emotionally charged versus original versions}\yccr{, as illustrated in Figure \ref{fig:overview}. We examine human performance on both binary fallacy detection (i.e., identifying whether an argument is fallacious, regardless of the specific fallacy type) and multi-way fallacy detection (i.e., classifying fallacy types).}
Our contributions are twofold:

\begin{itemize}[topsep=2pt,itemsep=-1pt,leftmargin=*]
    \item Emotional Framing with LLMs (\S\ref{sec:frame}): We examine eight LLMs from four model families for their ability (with zero-shot prompting) to inject \ycr{emotional appeals to }six emotions into argumentative text  
    \ycr{using} four framing strategies, while preserving the original content and logical structure.
    Results suggest that LLMs generally succeed in  
    \yc{injecting emotional expressions into fallacious arguments}  
    without disrupting their reasoning, 
    though mismatches frequently arise between the intended target emotions and those expressed in the text. We identify the best 4 models and 3 framing strategies to create a dataset consisting of 1,000 arguments for the human study. 

    \item Human Perception Study (\S\ref{sec:human}): We empirically assess how humans respond to emotionally enriched arguments, focusing on their fallacy detection ability and convincingness judgments.
    Our main findings include: Although humans often do not perceive the emotions expressed in arguments by LLMs, 
    (i) \yccr{when exposed to LLM-framed emotional arguments, their fallacy detection performance 
    drops by 5.5\% in accuracy (relative) on the binary detection task and by 14.5\% in F1 (relative) on the fallacy type classification task.} However, (ii) 
    \ycr{the LLM-framed emotional }arguments are rated only slightly more convincing than the original ones. 
    \yc{On the other hand, (iii) humans do rate convincingness higher when they detect no fallacies in arguments.}
    When considering human \emph{perceived} emotions, we find that 
(iv) fear, sadness, and enjoyment 
correlate with \yct{significantly} higher convincingness, and (v) humans perform \yct{significantly} better in fallacy detection when perceiving the positive emotion enjoyment \ycr{than when perceiving the negative states sadness or fear; both sadness and fear also impair performance relative to the neutral state.} 
    
\end{itemize}
    
Our findings contribute to computational and cognitive perspectives on argumentation, call for integrating affective dimensions into reasoning and persuasion, and point to risks of AI-driven manipulation in fallacious reasoning.\footnote{Code + Data: \url{https://github.com/NL2G/EMCONA-UTN}}

\begin{table*}[!ht]
\small
\centering
\resizebox{\linewidth}{!}{
\noindent\begin{tabularx}{\textwidth}{ >{\hsize=.1\hsize}X  >{\hsize=.2\hsize}X | >{\hsize=.7\hsize}X }
\toprule
\textbf{Strategy} & \textbf{Description}                            &  \textbf{Example}                \\ \midrule
Storytelling      & Sharing personal stories or anecdotes that evoke emotion.    & 
%Cell phones, high-tech devices that cause severe distractions and cyberbullying should be banned in schools\textbf{\incor{, because my friend's sister was involved in a tragic accident while texting and driving after leaving the school campus.}}              
Paranormal activity is real because I have experienced what can only be described as paranormal activity. \textbf{\incor{My friends and I had a thrilling adventure when we explored the reportedly haunted hotel and captured some compelling evidence.}}
\\ \midrule
Imagery           & Painting mental pictures to stir emotions.   & An increasing number of people are keeping ferrets as pets, so they must make wonderful companion animals. \textbf{\incor{Imagine the joy of watching them playfully dart around the room, their little bodies twisting and turning with pure delight.}}                                  \\ \midrule
Vivid Language    & Using descriptive, sensory, or emotionally charged words.    & I heard that the Catholic Church was \textbf{\incor{complicit} \cor{(involved)}} in a \textbf{\incor{horrific}} sex scandal cover-up\textbf{\incor{, allowing rot to fester in its hallowed halls.}} Therefore, my 102-year-old Catholic neighbor, who regularly enters that \textbf{\incor{tainted sanctuary} \cor{(Church)}}, is \textbf{\incor{equally stained by its moral decay} \cor{(guilty as well)}}!
                 \\  \midrule
Loaded Words      & Choosing words with strong positive or negative connotations.      & Being \textbf{\incor{sadly}} overweight leads to a \textbf{\incor{heartbreakingly}} shortened lifespan because it's \textbf{\incor{tragically}} unhealthy. \newline
People who \textbf{\incor{hate} \cor{(don't like)}} K-pop are racist.\\ \bottomrule        
\end{tabularx}}
\caption{Four strategies used to introduce emotional appeal. Texts in \incor{green} are the ones added for emotional framing by LLMs, and those in \cor{red} and brackets are the originals replaced.}\label{tab:strategies}
\end{table*}
\section{Related Work}
This work connects to (1) emotions in reasoning \& argumentation and (2) logical fallacies.

\paragraph{Emotions in Reasoning \& Argumentation}
Fallacy detection is a logical reasoning task that requires analyzing arguments beyond their surface meaning, focusing on how reasoning is structured and whether conclusions follow from premises. Evaluating convincingness is also tied to reasoning, as it involves judging the sufficiency and relevance of premises to conclusions \citep{wachsmuth-etal-2017-computational,mirzakhmedova2024large}. Both tasks, however, can be influenced by 
\ycr{distorting factors }arising from emotions, cognitive biases, and other subjective factors.

Prior work has shown that emotions systematically shape \textbf{human reasoning}, with particularly strong effects in deductive contexts. For instance, \citet{blanchette2006effect} demonstrate that emotional content can impair accuracy in conditional reasoning, while later work suggests that emotional relevance may sometimes facilitate logical evaluation \citep{blanchette2014does}. Emotions also bias probabilistic reasoning 
\citep{lerner2001fear}, and affective valence influences judgments  
\citep{slovic2005affect}. At a metacognitive level, mood can function as an implicit cue for confidence in one’s judgments \citep{schwarz1983mood}. Studies have also shown that positive emotions can broaden attentional scope \citep{basso1996mood,fredrickson2004broaden,jung2014emotions}, enhance critical thinking \citep{jung2014emotions}, and increase openness to others’ viewpoints \citep{isen2000some,tsai2023trait},
thereby potentially influencing argument evaluation. 
\ycr{Some research also reports the opposite view: negative emotions boost critical thinking, while positive ones impair it \citep{isen1983influence,mackie1989processing}.}

On the other hand, 
emotions have long been shown to influence 
\textbf{argument convincingness} in philosophy and psychology \citep[e.g.][]{kennedy1991theory,konat2024pathos,benlamine2015emotions}, yet in NLP they are often treated as a secondary factor \citep{greschner2024fearfulfalconsangryllamas,chen-eger-2025-emotions}: Research finds emotion to be a contributor to convincingness 
  \yc{through} annotators’ explanations of their judgments \citep{habernal-gurevych-2016-argument},  
\yc{the inclusion of the emotional appeal layer} (as one among many layers) in an argumentation model \citep{habernal2017argumentation}, or \yc{the integration of emotional appeal criterion} 
into an argument quality assessment model  \citep{wachsmuth-etal-2017-computational,ziegenbein-etal-2023-modeling}. 
\ycr{Three} recent works \yc{in NLP} place greater emphasis on emotion in argumentation. 
\citet{greschner2024fearfulfalconsangryllamas} annotate discrete emotions (e.g., joy, sadness) alongside convincingness for German arguments, finding that joy and pride correlate with higher convincingness, while anger lowers it. 
\citet{chen-eger-2025-emotions}  
find that emotions have no effect in more than half of the cases, but when an effect occurs, it is more likely to be positive, using GPT-4o to insert or remove emotions for counterpart argument generation.  However, their analysis treats emotion as a single broad category without distinguishing types. 
\ycr{\citet{greschner2025trustmeiconvince} propose a Contextualized Argument Appraisal Framework  
\yct{and show that convincingness rises with positive emotions and falls with negative ones, with argument familiarity the primary driver for emotional response.}}
\yc{ 
\ycr{To our knowledge, }no prior work in NLP  
has examined  
\yct{the} emotional effects \yct{on argument convincingness}  
\yct{in} the presence of fallacies.}

\paragraph{Logical Fallac\senew{ies}}
The study of logical fallacies dates back to Aristotle, who formalized them as flaws in reasoning that undermine the progression from premises to conclusion \citep{kennedy1991theory}. In NLP, multiple datasets 
have been collected for automatic fallacy detection across domains such as academia \citep{glockner-etal-2024-missci,glockner-etal-2025-grounding}, political debates \citep{goffredo2022fallacious,goffredo-etal-2023-argument}, news \citep{da-san-martino-etal-2019-fine,musi2022developing}, and social media/online forum\yct{s} 
\citep{sheng-etal-2021-nice,sahai-etal-2021-breaking,habernal-etal-2017-argotario,habernal-etal-2018-name,ramponi-etal-2025-fine,MACAGNO2022108501,macagno2022argumentation}.
Recently,   
\yct{fallacy detection} itself has also received renewed attention in the era of LLMs \citep{glockner-etal-2024-missci,jeong-etal-2025-large,cantin-larumbe-chust-vendrell-2025-argumentative}. 
%{glockner-etal-2024-missci,pan-etal-2024-llms,robbani-etal-2024-flee,helwe-etal-2024-mafalda,lei-huang-2024-boosting,jeong-etal-2025-large,cantin-larumbe-chust-vendrell-2025-argumentative}. 

Despite this growing body of work, only one study to date has evaluated human performance on \ycr{these} fallacy detection \ycr{datasets} \citep{helwe-etal-2024-mafalda}, 
based on just 20 samples, leaving open questions about how reliably humans can identify fallacies \ycr{under the same evaluation protocol 
for AI systems}. \ycr{This is important because establishing a human baseline is crucial for accurately judging AI performance.}  
Moreover, the emotional dimension of fallacious arguments has been largely neglected, as emotional appeal is often treated as a fallacy type in the community \citep[e.g.][]{jin-etal-2022-logical,musi2022developing,helwe-etal-2024-mafalda}.
%{jin-etal-2022-logical,helwe-etal-2024-mafalda,habernal-etal-2018-name,sahai-etal-2021-breaking,goffredo2022fallacious,musi2022developing,da-san-martino-etal-2019-fine,habernal-etal-2017-argotario}. 
\citet{MACAGNO2022108501} may be the only work that considers emotion as a separate dimension from fallacy, analyzing both fallacies and emotions in the tweets of four US presidents, though focusing solely on the textual use of emotive language.  

In this work, we create the first dataset for studying the interactions between fallacies, emotions, and convincingness in arguments, with the assistance of AI (\S\ref{sec:frame}). 
We then conduct a human study on this dataset and provide a detailed analysis based on human perceptions (\S\ref{sec:human}).

\section{AI-driven Emotional Framing}\label{sec:frame}
\ycr{ 
To identify optimal approaches (LLMs and emotional framing strategies) for generating the human-study data (\S\ref{sec:human}), we test whether LLMs can inject targeted emotional appeals into arguments while preserving the original content, especially the underlying fallacies. In the following, we describe our framing procedure, datasets and models, and evaluate LLM performance on this task.}

\paragraph{Emotional Framing} 
We consider \textbf{the six basic emotions} proposed by \citet{ekman1992argument}, namely anger, disgust, fear, enjoyment,\footnote{\ycr{Although `joy' is more commonly used in the NLP community, we adopt the term `enjoyment', as originally used by \citet{ekman1992argument}.}}  
sadness, and surprise. 
In addition, we prompt LLMs to introduce emotional appeal through \textbf{four distinct strategies}: storytelling, imagery, vivid language, and loaded words, as shown in Table~\ref{tab:strategies}. 

Images have been shown to elicit emotions \citep{gutierrez2025elicited,povskus2019effects}; here, we aim to achieve a similar effect in text by painting mental pictures with words, a strategy we refer to as \textbf{imagery}. We also adopt \textbf{storytelling} as an emotional framing strategy, motivated by the active research on the relationship between storytelling and emotions \citep[e.g.][]{mori2022computational,lugmayr2017serious}. In our setting, this is implemented by instructing models to share personal stories or anecdotes. \textbf{Loaded words} refers to the use of words with strong positive or negative connotations, whereas \textbf{vivid language} involves using descriptive, sensory, or emotionally charged expressions.

Besides differing in emotion  
mechanisms, we expect the four strategies to vary in the extent of modification. For storytelling and imagery, we prompt LLMs to add a sentence to the original argument, whereas vivid language and loaded words involve only word-level changes, ideally preserving a length \yc{similar to or slightly larger than} the original. 
The prompts used are provided in Appendix~\ref{app:emo_prompts}.

\begin{table}[!t]
\centering
\setlength\tabcolsep{1.5pt} 
\resizebox{1\linewidth}{!}{
\begin{tabular}{@{}llcc@{}}
\toprule
\textbf{Model Family} & \textbf{Checkpoint}                           & \textbf{Size} & \textbf{Reasoning?} \\ \midrule
OpenAI & gpt-4o                    & -   & \XSolidBrush   \\
\citep{openai2024gpt4technicalreport}             & gpt-4o-mini                          & -   & \XSolidBrush \\
             
             & o3-mini                    & -   & \Checkmark \\ \midrule
Llama3%\footnote{\url{https://huggingface.co/meta-llama}}
& llama-3.3-70b-instruct     & 70B  & \XSolidBrush  \\
\citep{grattafiori2024llama3herdmodels}             &  \\
              \midrule
Qwen3%\footnote{\url{https://huggingface.co/Qwen}} 
& qwen3-32b                    & 32B & \Checkmark \\
\citep{yang2025qwen3technicalreport}              & qwen3-8b                    & 8B & \Checkmark \\  \midrule
DeepSeek-R1  & deepseek-r1-distill-llama-70b     & 70B  & \Checkmark \\ 
\citet{deepseekai2025deepseekr1incentivizingreasoningcapability} & deepseek-r1-0528-qwen3-8b                    & 8B & \Checkmark \\
\bottomrule
\end{tabular}}
\caption{LLMs used in this work. We use \texttt{ds-model} as an abbreviation for DeepSeek-distilled models in the text.
}\label{tab:llms}
\vspace{-.3cm}
\end{table}

\paragraph{LLMs}
We evaluate eight LLMs from four model families, encompassing both open-source and commercial models, as well as reasoning and non-reasoning models, as summarized in Table~\ref{tab:llms}. For DeepSeek-R1, we select the distilled models for LLaMA3.3 and Qwen3 as counterparts. 
All models are accessed via the OpenRouter API.\footnote{\url{https://openrouter.ai/}} We set the temperature to 0.6 for all models and tasks, keeping other parameters at their default values.

\paragraph{Dataset}
We use the \data{} dataset from \citet{jin-etal-2022-logical}, which contains approximately 2.5k fallacious arguments spanning 13 types of logical fallacies\ycr{, including emotional appeal. } 
The arguments were collected from online educational resources on logical fallacies. 
\yc{While some fallacy datasets are fully synthetic \citep[e.g.][]{mouchel2024logicalfallacyinformedframeworkargument,li-etal-2024-reason} and some are derived from  
long articles, speeches, or discourses \citep[e.g.][]{goffredo-etal-2023-argument,glockner-etal-2024-missci,sheng-etal-2021-nice}, we select this dataset because it consists of human-written arguments that are comparatively short, standalone (i.e., context-independent), yet exhibit clear premise–conclusion structures. Given that evaluating fallacies, emotions, and convincingness is inherently complex and cognitively demanding, beginning with a simpler dataset makes the study more manageable. 
} 
We \senew{thus} focus on \ycr{the five fallacy types most frequent in the dataset}---\emph{Faulty Generalization, Ad Hominem, Ad Populum, False Causality} and \emph{Circular Claim}---\yc{also} to reduce the cognitive load on annotators.  
We refer  
to  
\yc{Appendix~\ref{pdf:fallacy}} for detailed definitions \ycr{and examples} of these fallacies. After removing duplicated instances and those that explicitly mention the fallacy type,\footnote{E.g., ``This is an example of a circular claim.''}
1,245 unique arguments remain in our dataset. 

\yc{We then use the prompt-based classifier from \citet{chen-eger-2025-emotions} with gpt-4o and qwen3-32b to filter out non-standalone arguments and automatically 
\yct{generate} short claims for arguments.  
\yct{During the convincingness evaluation in \S\ref{sec:human}, claims are displayed alongside the arguments to serve as an anchor.}
Arguments classified as standalone by both models are retained, resulting in \textbf{538 arguments}. For each retained argument, we randomly select the claim generated by either gpt-4o or qwen3-32b.  
\yct{In Appendix~\ref{app:arg}, we show that the classifier, when implemented with both models, achieves 100\% accuracy on our dataset.}
}

\paragraph{Evaluation}
We consider \textbf{5 evaluation dimensions} for this task: \ycr{(1) Reasoning preservation:  
\yct{We test whether the fallacy remains constant across the original and emotionally enriched versions to fairly compare human fallacy detection on the original vs.\ emotional arguments.} 
(2) Coherence: We assess textual coherence (as in general text generation evaluation \citep{10.1162/tacl_a_00373,zhang2025litransproqallmbasedliterarytranslation}) and emotional coherence, i.e., whether the injected tone is natural and consistent. 
(3) Length difference: We compare argument length to verify expected changes under different strategies. 
(4) Emotional appeal: We compare generated vs. original emotional-appeal levels to verify that the injection functions as intended. 
(5) Emotion match: \yct{We check} whether the expressed emotion matches the target specified at generation time. 
We refer to Appendix~\ref{app:screenshots} for the detailed definitions of these dimensions in annotation guidelines presented to annotators. }

We randomly sample 5 distinct arguments per fallacy type \ycr{(5 $\times$ 5 fallacy types = 25 originals)} from the filtered \data{} dataset and run each model on them. This results in 3,840 generated arguments.\footnote{We began with 25 original arguments $\times$ 8 models $\times$ 6 emotion types $\times$ 4 strategies, totaling 4.8k generated arguments. However, in this experiment, we applied the argument classifier using the same model as used for generation. As a result, some original arguments were filtered out \yc{before generation}, resulting in fewer than 4.8k generated arguments.}
For \textbf{human evaluation}, we randomly sample 25 arguments per generation model from the 3.8k generated samples, totaling 200 arguments.
\ycr{We also annotate the original arguments for coherence and emotional appeal to establish baselines against which to compare generated arguments.
 Annotators disagreed on only 2 and 1 out of 20 overlapping annotation instances for reasoning preservation and coherence, respectively, demonstrating decent levels of annotation agreements. The emotion related dimensions received lower agreements, with a Pearson correlation of 0.2 for emotional appeal and a Cohen's Kappa of 0.306 for emotion match on the full samples. Nevertheless, using an independent group of annotators (see \S\ref{sec:limit}), we obtained largely comparable conclusions for these criteria, indicating good system-level reproducibility. See Appendix~\ref{app:frame_annon} for the annotation details.}

\ycr{
\yct{For brevity, full \textbf{evaluation results} appear in Appendix~\ref{app:frame_eval}, and we present a condensed summary here: } 
Overall, LLMs increase the emotional appeal of arguments in 94\% of cases while preserving the underlying fallacious reasoning in 89\%, although intended emotions diverge from human judgments in 74\% of cases. LLMs also produce near-human-level coherence (0.88–0.98 vs.\ 1.00), with o3-mini the only exception (0.60).  
Based on these results, we select the strategies and models used to generate the human-study data in \S\ref{sec:human}, as follows:} 
\begin{itemize}[topsep=2pt,itemsep=-1pt,leftmargin=*]
    \item \textbf{Selected Strategies: } 
Different emotional framing strategies lead to variations in the generated arguments across these dimensions. For example, imagery and storytelling tend to produce longer arguments, preserve the original reasoning structure well, and evoke stronger emotions, but they do so at the cost of reduced coherence---particularly in the case of imagery. Vivid language produces the most coherent texts; however, it often disrupts the original reasoning structure of the arguments. 
Since our goal is to create data to study the effects of emotions on fallacy detection, we must first ensure that emotional framing does not compromise the logical structure of the arguments. Therefore, \textbf{we exclude vivid language from further experiments}. 

\item \textbf{Selected Models: }
Similar to our selection of strategies, we choose the best models primarily based on their ability in reasoning preservation, while ensuring they do not perform substantially worse than others in coherence (e.g., o3-mini, with an average coherence score of only 0.60). Since the emotion match rates are generally low, in \S\ref{sec:human}, we do not base our analysis on the emotions assigned to LLMs, but instead ask human annotators to select the emotions they perceive. 
Based on these criteria, we select the best model from each model family to diversify data points for the human study:
\textbf{qwen3-32b, ds-llama-70b, gpt-4o-mini}, and \textbf{llama-3.3-70b} as the \textbf{top 4 models}.
\end{itemize}
\section{Human Fallacy Detection}\label{sec:human}

\begin{figure*}[!ht]
    \centering
    \includegraphics[width=0.36\linewidth]{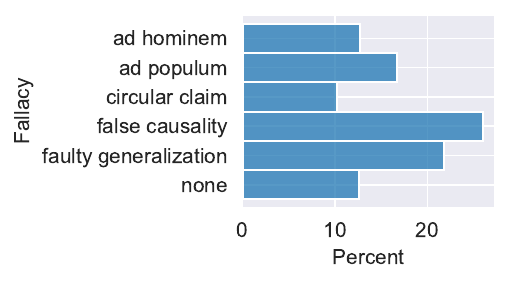}
    \includegraphics[width=0.37\linewidth]{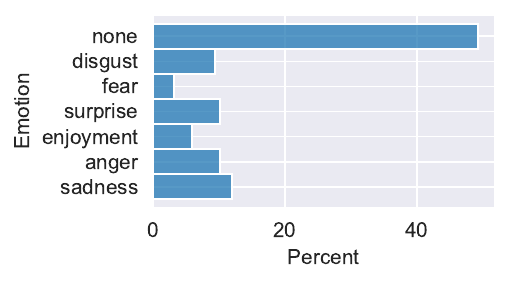}
    \includegraphics[width=0.25\linewidth]{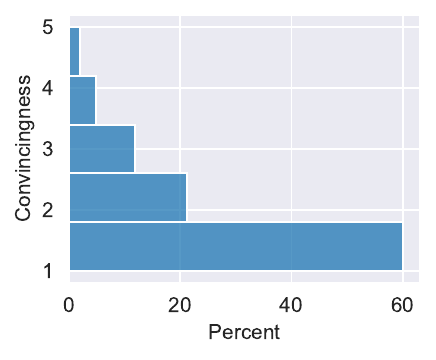}
    \vspace{-.4cm}
    \caption{Label distributions: percentages of each label.}
    \label{fig:label_dis}
    \vspace{-.5cm}
\end{figure*}

\begin{table}[]
\centering
\small
\resizebox{\linewidth}{!}{
\begin{tabular}{@{}lc|lc|lc@{}}
\toprule
\multicolumn{2}{l|}{\textbf{Emotion}} & \multicolumn{2}{l|}{\textbf{Strategy}} & \multicolumn{2}{l}{\textbf{Model}} \\ \midrule
original & \multicolumn{5}{c}{200} \\ \midrule
enjoyment        & 133       & loaded words      & 274      & qwen3-32b        & 214    \\
anger            & 135       & imagery           & 264      & ds--llama-70b    & 188    \\
fear             & 139       & storytelling      & 262      & gpt-4o-mini      & 194     \\
surprise         &122       &                   &          & llama-3.3-70b    & 204     \\
sadness          & 145       &                   &          &                  &        \\
disgust          & 126       &                   &          &                  &        \\ \midrule
\textbf{Total}            & \multicolumn{5}{c}{1,000}                                           \\ \bottomrule
\end{tabular}}
\caption{Statistics of the human study data, including the number of arguments for each target emotion, emotional framing strategy, and generation model.}\label{tab:stats}
\vspace{-.5cm}
\end{table}

\paragraph{Data}
We randomly sample 40 arguments per fallacy type from \data{}. For each argument, we generate 4 counterpart arguments by randomly using two models, two emotional framing strategies, and one emotion type. In the end, our data contains 1,000 arguments (= 40 arguments $\times$ 5 fallacies $\times$ 2 models $\times$ 2 strategies $\times$ 1 emotion + 200 original ones); the statistics are presented in Table~\ref{tab:stats}. We randomly split the 1k arguments into 20 batches of 50 annotation instances each.

\paragraph{Annotators}
We recruit annotators from the crowdsourcing platform Prolific.\footnote{\url{https://www.prolific.com/}} 
We use Prolific’s pre-screening function to recruit annotators based in the US or UK who are native English speakers and have completed higher education (at least a technical/community college degree). Each batch is annotated by the same group of three annotators. The total cost is 360 Euros. 

\paragraph{Annotation Criteria} 
Annotators 
are instructed to annotate each argument according to \textbf{three criteria}.
(1) \textbf{Emotion type}: in contrast to \S\ref{sec:frame}, the question is framed  
subjectively here: `Which emotion do you feel strongest?', while using the same label set.
(2) \textbf{Fallacy type}: select one of the five specified fallacies, or `none' if no fallacy is detected or if the identified fallacy is not among the listed types. 
\yc{Before annotation, annotators completed a quiz with five simple fallacy detection tasks taken from the examples in \citet{jin-etal-2022-logical}, achieving an average score of 3.81/5. \ycr{This score is much higher than the random-choice baseline for a six-way classification setup across five tasks ($\frac{1}{6}\times5\approx0.83$), indicating the overall reliability of the annotators on this task. }}
(3) \textbf{Convincingness}: rate on a Likert scale of 1-5 given the claim, where 1 denotes `not convincing at all' and 5 is `extremely convincing'; if the claim does not fit the argument, select `N/A'. We refer to Appendix~\ref{app:screenshots} for screenshots of the annotation interface, guidelines\ycr{, and the tasks of the fallacy detection quiz}. 

\paragraph{Label Distribution}
Figure~\ref{fig:label_dis} shows the label distributions for each annotation task. $\sim$87\% of arguments were annotated as fallacious, with false causality selected most frequently and circular claim least frequently. Only 51.6\% of arguments were annotated with an emotion, with sadness being the most common and fear the least, suggesting that emotions expressed by LLMs are often not perceived by humans. This reflects the gap between emotional expression and perception\yc{, i.e., writer’s/text’s vs.\ reader’s perspectives \citep{5286061,haider-etal-2020-po,wang2025emotion,bostan-etal-2020-goodnewseveryone}}. 
As for \textbf{convincingness}, only 49 out of 3,000 labels (1.6\%) are `N/A', indicating that the automatic claim generation process is largely reliable. Most arguments were rated as ``1 Not convincing at all'' (59.1\%), with the number of ratings decreasing as the score increases; the average rating is 1.67. This distribution indicates the general unconvincingness of fallacious arguments.

\paragraph{Annotation Agreement \& Aggregation}
We acknowledge the subjectivity involved in annotating perceived convincingness and emotions in arguments \citep{wachsmuth-etal-2017-computational, chen-eger-2025-emotions}, as well as the variability in human logical reasoning abilities. Nonetheless, we report full inter-annotator agreement results in Appendix~\ref{app:iaa}, including Cohen’s Kappa \citep{cohen1960coefficient}, Pearson correlation \citep{pearson1895vii}, and majority/full agreement metrics \citep{wachsmuth-etal-2017-computational}. In general, for classification tasks, the more logic-driven criterion, fallacy, 
achieves higher agreement than emotion  
(e.g.\ Kappa: 0.014--0.591 vs.\ -0.005--0.139). 
The average majority agreement \yc{(i.e., percentage of instances with at least 2 identical labels)} across annotation batches ranges from 73\% (emotion) to 80.5\% (convincingness); 
the convincingness criterion achieves the highest agreement mainly because the majority of the ratings are 1.

We  
adopt \textbf{a single aggregated label} per argument as the gold standard to evaluate \textbf{the effects of emotional framing} introduced by LLMs. To do so, 
we aggregate the discrete labels for emotion  
\yct{or} fallacy using \emph{majority voting}: if at least two \yct{out of three} annotators select the same label, it is taken as the final label. 
In cases where no majority is reached, we adopt the label from the annotator who agree\se{s} most with the others on average. 
For convincingness, \yct{to account for differences in rating behavior, we normalize each annotator’s convincingness scores using \emph{z-scores}, ensuring comparability across annotators and batches \citep{ma-etal-2018-results,ma-etal-2019-results};} 
we then average the z-score normalized convincingness scores over annotators to obtain the final score for each argument.
In contrast, 
we analyze the relationship between \textbf{perceived} emotions, convincingness, and \yct{the fallacy detection performance} by considering \textbf{individual annotations} from each annotator and reporting averages\yct{; we also apply z-score normalization to convincingness for the comparability across different annotators.}

\paragraph{Effects of AI-driven Emotional Framing}

\begin{figure}
    \centering
    \includegraphics[width=.8\linewidth]{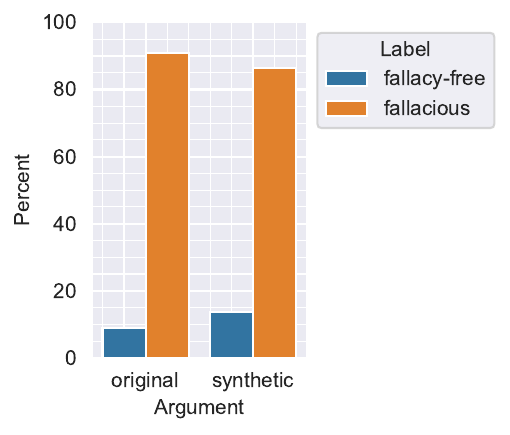}
    \caption{Human-perceived fallacy distribution (fallacy-free vs. fallacious) for original and synthetic emotionally framed arguments using all emotional framing strategies.}
    \label{fig:fallacy_dist_comp}
\end{figure}

\begin{table}[!t]
\centering
\small
\setlength{\tabcolsep}{1.7pt}
\resizebox{\linewidth}{!}{
\begin{tabular}{lccc|ccc}
\toprule
& \multicolumn{3}{c|}{Fallacy Detection} & \multicolumn{3}{c}{Convincingness} \\
model                         & imagery & loaded & story & imagery & loaded & story \\ \midrule
ori                           & \multicolumn{3}{c|}{0.563}             & \multicolumn{3}{c}{-0.031}            \\ \midrule
ds-llama-70b & 0.533   & 0.524        & 0.530        & 0.099   & -0.029       & 0.233        \\
gpt-4o-mini                   & 0.451   & 0.520        & 0.444        & -0.021  & -0.076       & -0.052       \\
llama-3.3-70b        & 0.530   & 0.466        & 0.474        & 0.068   & -0.115       & 0.105        \\
qwen3-32b                     & 0.504   & 0.380        & 0.423        & -0.097  & -0.069       & 0.099  \\ \midrule
AVG & 0.505 & 0.473 & 0.468 & 0.012 & -0.072 & 0.096 \\ \bottomrule     
\end{tabular}}
\caption{Macro F1 in fallacy detection and average convincingness scores (z-score normalized) per model and strategy.}\label{tab:ai}
\vspace{-.5cm}
\end{table}

\yccr{
As Figure~\ref{fig:fallacy_dist_comp} shows, we compare human-perceived fallacy distributions between the original and synthetic arguments using all emotional framing strategies. The proportion of arguments labeled as fallacious (i.e., not `none') decreases from $\sim$91\% to $\sim$86\% (a 5.5\% relative drop), suggesting that LLM-based emotional framing makes fallacious arguments harder for humans to detect overall.} 
In Table~\ref{tab:ai}, 
we report the macro F1 scores for human fallacy detection \yc{(left)} and the normalized convincingness ratings \yc{(right)} for each model and emotional framing strategy, using aggregated labels. 
\yc{As Table~\ref{tab:ai} (left) presents}, humans consistently perform worse on LLM-generated arguments than on original ones, 
with an average \yccr{relative} drop of $\sim$14.5\% (0.380–0.533 vs.\ 0.563). 
Among all combinations, qwen3-32b using the loaded words strategy has the strongest negative impact on human reasoning (0.380), while ds-llama-70b with imagery has the least impact (0.533). On average, qwen3-32b impairs reasoning the most (0.436),  
and loaded words appears to be the most detrimental strategy for 3 out of 4 models (except gpt-4o-mini). The least impactful model and strategy are ds-llama-70b and imagery, respectively.

One might question whether longer arguments make fallacies harder to detect, since the relative span of the fallacious content becomes smaller.  
We compute the Spearman correlation between length difference and performance difference,\footnote{Specifically, we group arguments into bins of 5 tokens by length difference, then correlate the average length difference with the change in F1 scores across these bins. 
We analyze bins with length differences between 0 and 40 tokens, which already account for 97\% of the data.}  
finding a moderate to strong positive correlation between the number of added tokens and \yc{fallacy} performance improvement ($\rho = 0.619$). 
This observation  
actually   
\yc{aligns} with \yc{above:}  
loaded words, the strategy that adds the fewest tokens, leads to the lowest fallacy detection performance. 
\yc{One possible reason is that, as shown in Table~\ref{tab:strategies}, imagery and storytelling typically preserve the original fallacious arguments while adding extra content, whereas loaded words insert tokens directly into the text, making the fallacious content more diffuse.}

As for \textbf{convincingness}, 
all scores are close to the mean, with z-scores ranging from –0.115 to 0.233. LLM-generated arguments are rated slightly higher than original ones, with an average increase of 0.04 standard deviations. The most convincing arguments are generated by ds-llama-70b using storytelling, while llama-3.3-70b with loaded words is rated least convincing. Overall, ds-llama-70b produces the most convincing arguments (0.101), and storytelling emerges as the most effective persuasion strategy (0.096). The least convincing model and strategy are gpt-4o-mini and loaded words, respectively.

\begin{figure}[!t]
    \centering
    \includegraphics[width=\linewidth]{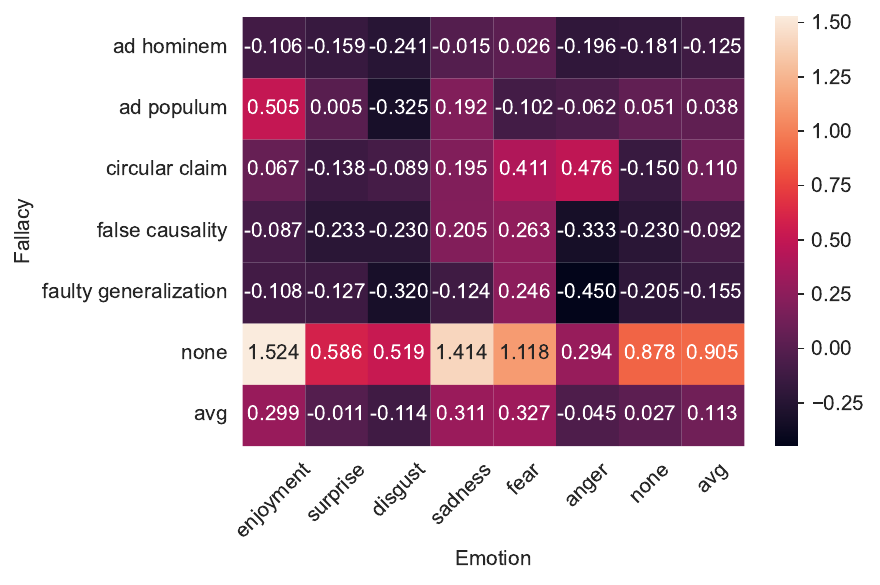}
    %\vspace{-.3cm}
    \caption{Average convincingness scores (z-score normalized) for each emotion and fallacy category based on human annotations \yct{\textbf{(perceived convincingness, emotions, and fallacies)}}. Full pairwise t-test results appear in Tables~\ref{tab:ttest_conv_fallacy} and ~\ref{tab:ttest_conv_emo} (Appendix~\ref{app:ttest}).}
    \label{fig:conv_matrix}
    %\vspace{-.5cm}
\end{figure}

\paragraph{Perceived emotions/fallacies vs.\ convincingness}
Figure~\ref{fig:conv_matrix} shows the average convincingness scores for each emotion and fallacy category; here, the categories were assigned by the annotators. 
\ycr{We report t-test results across emotion and fallacy categories in Tables~\ref{tab:ttest_conv_fallacy} and~\ref{tab:ttest_conv_emo} (Appendix~\ref{app:ttest}).} 
We observe that arguments identified as fallacy-free (`none') consistently receive above-average scores across different emotions, ranking first among fallacy categories on average (0.905), \ycr{significantly} ahead of the second- and third-most convincing categories, circular claim (0.110; $p\leq0.01$) and ad populum (0.038; $p\leq0.01$). The remaining fallacies receive below-average ratings, with faulty generalization being rated least convincing ($-0.155$). Among emotions, fear, enjoyment, and sadness are rated as the most convincing on average (0.299–0.327)\ycr{, significantly more convincing than the other categories, including the non-emotional one ($p\leq0.01$); }  
all of the other emotions receive around-average ratings, ranging from –0.114 for disgust to 0.027 for no emotion (`none')\ycr{, where arguments with anger and disgust are rated significantly less convincing than non-emotional ones ($p\leq0.01$).}

\textbf{Overall}, the results suggest that humans perceive fallacies as flaws in persuasion, leading to lower convincingness scores. When positive emotions like enjoyment are perceived, people 
\yct{may} 
show greater openness to others' viewpoints \citep{isen2000some,tsai2023trait}, resulting in higher convincingness ratings. Fear and sadness also appear to enhance convincingness.

\begin{table}[!t]
\centering
\resizebox{\linewidth}{!}{
\iffalse
\begin{tabular}{@{}lcc@{}}
\toprule
          & F1  & \(p\)  \\ \midrule
enjoyment & \textbf{0.528} & 0.156 \\
surprise  & 0.464 & 0.600\\
anger     & 0.501 & 0.214\\
disgust   & 0.461 & 0.663\\
sadness   & 0.337 & 0.054\\
fear      & \underline{0.283} & 0.023\\ \midrule
none      & 0.437 & - \\ \bottomrule
\end{tabular}\fi
\begin{tabular}{@{}ccccccc|c@{}}
\toprule
 & enjoyment & surprise  & anger & disgust &sadness & fear &  none  \\ \midrule
$F1$  & \textbf{0.528} & 0.464 & 0.501 & 0.461 & 0.337 & \underline{0.283} & 0.437 \\
\(p\) & 0.156 & 0.600 & 0.215 & 0.663 & 0.054 & 0.023 & - \\ \bottomrule
\end{tabular}}
\caption{Macro-F1 for fallacy detection by \textbf{perceived emotion} (averaged over annotators); the best value is in bold; the worst is underlined. \(p\)-values from t-tests comparing each emotion to the ‘none’ category are shown in the last column; full pairwise t-test results appear in Table~\ref{tab:ttest_f1_emo} (Appendix~\ref{app:ttest}).}\label{tab:f1_emo}
\vspace{-.5cm}
\end{table}
\vspace{-.2cm}
\paragraph{Perceived emotions vs.\ fallacy detection}
Table~\ref{tab:f1_emo} displays the macro F1 scores of humans in detecting fallacies, averaged across annotators, for each \yc{annotator-assigned} emotion category. \ycr{We report the full significance test results in Table~\ref{tab:ttest_f1_emo} in Appendix~\ref{app:ttest}.}
We observe that 
\ycr{human performs best when perceiving enjoyment, significantly better than when perceiving sadness and fear (0.528 vs.\ 0.337/0.283; $p = 0.016/0.012$). Fear and sadness appear to impair fallacy detection relative to the neutral condition (0.283/0.337 vs.\ 0.437; $p = 0.023/0.054$).}
Positive emotions have been shown to broaden attentional scope \citep{derryberry1994motivating,basso1996mood,fredrickson2004broaden} and enhance critical thinking \citep{jung2014emotions} compared to negative emotions, which aligns with our observations. However, we do not find that a neutral emotional state consistently\ycr{/significantly} leads to better reasoning performance than emotions, as observed in \citet{jung2014emotions}. 

Interestingly, the emotion that most enhances reasoning ability (enjoyment) and the two that most impair it (fear and sadness) are also the most convincing emotions observed in Figure~\ref{fig:conv_matrix}. This may suggest \textbf{two pathways to perceived convincingness of fallacious arguments}: (1) positive emotions like enjoyment may increase openness to persuasion while still supporting logical scrutiny, enabling people to detect fallacies 
yet remain receptive to \yct{the major claims of the} arguments; and (2) negative emotions like fear and sadness may impair reasoning processes, making it harder to recognize logical flaws, thus increasing perceived convincingness through reduced critical evaluation.

\paragraph{\ycr{Key Findings}}
\yc{
Exposure to LLM-framed arguments reduces fallacy detection performance compared to originals, most notably with loaded words, but only slightly increases perceived convincingness. Convincingness ratings are higher when there are no errors detected. Emotion-wise: perceived enjoyment improves fallacy detection relative to 
negative states fear and sadness, whereas fear and sadness impair human reasoning; the three emotions also 
correlate with higher convincingness.}

%\input{sections/5_discussion}
%\vspace{-.1cm}
\section{Concluding Remarks}
%\vspace{-.2cm}
This work presents, to our knowledge, the first systematic study of how (AI-framed) emotions affect fallacy detection and argument convincingness. We benchmarked LLMs for their ability to inject emotions into arguments and created a dataset of emotionally framed fallacious arguments, which we then evaluated through human studies. Our results show that while LLMs 
\ycr{can frame textual emotions without disturbing the reasoning structure of arguments from premise to conclusion}, humans often do not perceive the \ycr{intended emotions}, leading to a gap between emotional expression and subjective experience. Importantly, 
\yct{AI-driven} emotional framing reduces human performance in fallacy detection but only modestly increases convincingness. Certain perceived emotions, such as fear, sadness, and enjoyment, are associated with higher persuasive impact; enjoyment also facilitates fallacy detection, whereas sadness and fear impair it.

These findings highlight both opportunities and risks. On the one hand, our study provides new insights into the interplay between emotions and reasoning, offering resources for research in computational argumentation.
\yccr{Future work could also explore the effects of emotions on LLM behavior in argumentation. For example, \citet{payandeh-etal-2024-susceptible} find that LLMs are substantially more often convinced when presented with logical fallacies in a multi-round debate setting, but do not consider the role of emotions.}
On the other hand, the ability of LLMs to strategically inject emotions into arguments underscores potential misuse for manipulation and misinformation. 
Future work should expand the scope to more diverse and real-world arguments, explore refined prompting strategies, and investigate mitigation methods to safeguard against malicious applications. 
%\yccr{\citet{payandeh-etal-2024-susceptible} find that LLMs are substantially more often convinced when presented with logical fallacies in a multi-round debate setting; future work could explore the effects of emotions on LLMs’ fallacy detection ability and convincingness judgments to assess whether LLMs exhibit human-like biases.}

\section*{Limitations \& Ethical Concerns}\label{sec:limit}
Limitations of this work include: (1) To  
\se{mitigate}
annotation complexity, we only leveraged a simple dataset based on online educational resources for logical fallacy detection, which may differ from real-world arguments in terms of length, topics, and complexity of logical structure etc. This may have biased the results and weakened the generalizability in real-world scenarios. Also,  
\yc{there are no genuine fallacy-free arguments in our dataset} to make a fairer comparison for convincingness between fallacy-free and fallacious arguments. (2) The evaluation of LLMs on the emotional framing task was conducted on a relatively small sample (25 outputs per model) and without repeated runs. This design may have introduced randomness into the results. 
(3) We did not systematically optimize prompts for emotional framing, but relied on a single heuristic prompt that appeared to yield reasonable outputs across models.  
Consequently, LLM performance in this task may have been underestimated. 
(4) We observe low agreements among crowdworkers for emotion-related annotations in \S\ref{sec:frame} (0.2 Pearson's $r$). However, using the two student annotators from \S\ref{sec:frame} yielded largely comparable results: synthetic arguments were still rated higher than originals in terms of emotional appeal (2.41 vs.\ 2.08 EA on average), ds-qwen3-8b produced the most emotionally appealing arguments (2.55) while o3-mini the least (2.18), and emotion match rates remained consistently low across models (23.2\%-52.9\%). 

Our findings show that the emotionally framed arguments by LLMs can impair the logical detection performance of humans, which raises risks of misuse for generating persuasive misinformation, propaganda, or manipulative content. Although our experiments were conducted under controlled research settings, the same methods could be applied to deliberately obscure logical flaws in harmful ways.

We used ChatGPT for text refin\se{ement} and search %ing 
for relevant papers. We obtained annotator consent (via Google Forms) to use their annotations.

\section*{Acknowledgements}
We thank the meta-reviewer and reviewers for their thoughtful and constructive comments, which have improved the final version of the paper.  
The NLLG Lab gratefully acknowledges support from 
%the Federal Ministry of Education and Research (BMBF) via the research grant “Metrics4NLG” and 
the German Research Foundation (DFG) via the Heisenberg Grant EG 375/5-1. This project has been conducted as part of the EMCONA (The Interplay of Emotions and Convincingness in Arguments) project, which is funded
by the DFG (projects EG-375/9-1 and KL2869/12–1).

\bibliography{custom}
%\bibliographystyle{acl_natbib}

%\onecolumn

\appendix
%\clearpage
\section{AI-driven Emotional Framing}\label{app:emo_prompts}
\subsection{Emotional Appeal Injection Prompt}
Table \ref{tab:prompt_frame} presents the prompts we used to inject emotional appeals into fallacious arguments.

\subsection{Argument Classifier}\label{app:arg}
\begin{figure}[!ht]
    \centering
    \includegraphics[width=\linewidth]{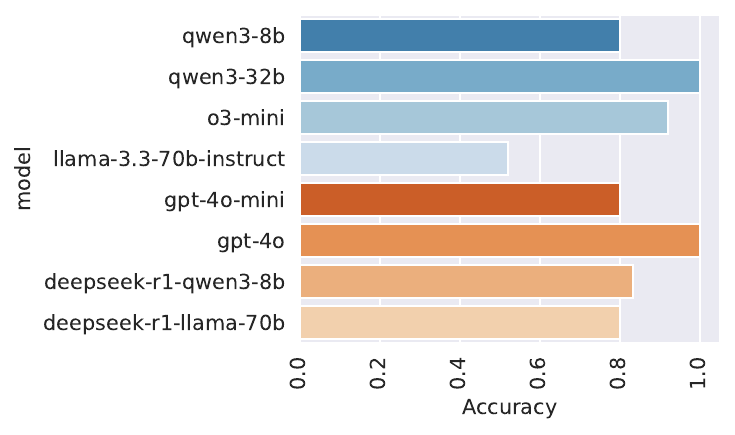}
    \caption{Accuracy of the prompt-based argument classifier from \citet{chen-eger-2025-emotions} using different LLMs on our dataset in both argument classification and claim generation.}
    \label{fig:classifier}
\end{figure}

We use the prompt-based classifier from \citet{chen-eger-2025-emotions} to filter out non-standalone arguments and automatically assign short claims to arguments.\footnote{Their prompt instructs the model to generate a claim when the text is identified as a standalone argument.} Since our text domain differs substantially from theirs (political debates), we first validate the reliability of the classifier using different models on 25 samples from our dataset.  
Two student annotators who are fluent in English, hired from German universities, annotated the 25 samples. \ycr{See Appendix~\ref{pdf:pilot} for the  annotation guideline for this criterion and Figure~\ref{fig:screenshot_sheet} for the annotation interface (Task 3 `Claim Correctness').}
As shown in Figure~\ref{fig:classifier}, qwen3-32b and gpt-4o perform best, both achieving perfect accuracy (1.0), whereas llama-3.3-70b reaches only about 0.5. 
\yct{This classifier performs better than reported in \citet{chen-eger-2025-emotions}. A very likely reason is that \citet{chen-eger-2025-emotions} evaluate on real-world political debates, whereas our short, human-curated arguments have clear evidence–claim structures with claims mostly explicit, making the task substantially easier for models.}

\subsection{Annotation}\label{app:frame_annon}
For the two more objective dimensions, namely reasoning preservation and coherence, we employed two student annotators from German universities, both fluent in English and enrolled in Master’s programs. They jointly labeled 20 overlapping instances for agreement checking and annotated the remaining instances independently. For the emotion-related dimensions, emotional appeal and emotion match, we recruited crowdworkers via Prolific,\footnote{\url{https://www.prolific.com/}} requiring them to be native English speakers based in the US or UK with a completed higher education degree. Each instance was annotated by three workers, and each worker evaluated 25 arguments. 
\ycr{See Appendix~\ref{app:screenshots} for the annotation guidelines and screenshots of the annotation interfaces. The annotation agreements are reported separately for each dimension in Appendix~\ref{app:frame_eval}.}

\subsection{Evaluation}\label{app:frame_eval}

% \usepackage{booktabs}
\begin{table*}[!ht]
\small
\centering
\setlength{\tabcolsep}{1.7pt}
\resizebox{\linewidth}{!}{
\begin{tabular}{@{}lccccc|ccccc|ccccc|ccccc|ccccc@{}}
\toprule
                       & \multicolumn{5}{c|}{\textbf{Imagery}}                                      & \multicolumn{5}{c|}{\textbf{Loaded Words}}                             & \multicolumn{5}{c|}{\textbf{Storytelling}}                              & \multicolumn{5}{c|}{\textbf{Vivid Language}}                               & \multicolumn{5}{c}{\textbf{AVG}}                                              \\
                       & RP            & Coh       & LD            & EA            & EM          & RP         & Coh       & LD           & EA            & EM          & RP         & Coh       & LD            & EA            & EM          & RP         & Coh          & LD            & EA            & EM          & RP            & Coh          & LD            & EA            & EM          \\ \midrule
ds-llama-70b$^{*}$  & \checkmark          & \checkmark       & \underline{14.4}    & .10          & \underline{.0}     & \checkmark       & \checkmark       & 6.3          & .19          & \checkmark          & .90       & .90       & \textbf{31.3} & .06  & 30.0          & .86       & \checkmark          & \textbf{14.5} & -.43          & 25.0          & .94          & \textbf{.98} & \textbf{16.6} & -.02          & 38.8 \\
ds-qwen3-8b   & \checkmark          & \checkmark       & 17.0          & \underline{-.49}          & 20.0          & \underline{.40} & \checkmark       & 3.7          & \textbf{.90}          & 60.0          & \checkmark       & \underline{.57} & \underline{21.8}    & .00          & \textbf{50.0} & \underline{.57} & .86          & 6.1           & \textbf{.39} & \textbf{71.4} & \underline{.74}    & .86          & \underline{12.2}    & \textbf{.20} & \textbf{50.4}          \\
gpt-4o                 & \checkmark          & .70       & 21.1          & \textbf{.45}          & 30.0          & \checkmark       & \checkmark       & 3.7          & -.09          & 33.3          & \underline{.80} & \checkmark       & 23.5          & \textbf{.50} & 80.0          & .75       & \checkmark          & 4.8           & -.18    & 25.0          & .89          & .93          & 13.3          & .17          & 42.1          \\
gpt-4o-mini$^{*}$            & \checkmark          & \checkmark       & 19.7          & .26          & 40.0          & \checkmark       & \checkmark       & 4.1          & -.16          & \underline{.0}     & \checkmark       & .86       & 23.0          & .07          & 42.9          & .60       & \checkmark          & 8.8           & .26          & 0.0          & .90          & .96          & 13.9          & .11          & \underline{20.7}          \\
llama-3.3-70b$^{*}$ & \checkmark          & .78       & 21.9          & .37          & 33.3          & \checkmark       & \checkmark       & 5.2          & -.15          & 75.0          & \checkmark       & \checkmark       & 23.0          & -.05          & 40.0          & .71       & \checkmark          & 11.7          & \underline{-.59}          & \underline{42.9}    & .93          & .94          & 15.4          & -.11          & 47.8          \\
o3-mini                & \underline{.89}    & \underline{.00} & 20.6          & -.08    & 10.0          & .75       & \underline{.75} & 3.7          & \underline{-.21}          & 25.0          & \checkmark       & .67       & 25.9          & -.14          & 0.0          & .80       & \checkmark          & \underline{4.5}     & -.25          & 60.0          & .86          & \underline{.60}    & 13.7          & \underline{-.17}    & 23.8          \\
qwen3-32b$^{*}$              & \checkmark          & .83       & 21.5          & .22          & \textbf{33.3} & \checkmark       & \checkmark       & \textbf{6.3} & -.22 & \textbf{83.3} & \checkmark       & \checkmark       & 22.2          & \underline{-.39}    & 0.0          & \checkmark       & \checkmark          & 8.8           & -.39          & 40.0          & \textbf{\checkmark} & .96          & 14.7          & .02          & 39.2 \\
qwen3-8b               & \checkmark          & .86       & \textbf{22.5} & \textbf{-.22} & 14.3          & .83       & .83       & \underline{3.5}    & .49    & 50.0          & .83       & \checkmark       & 24.9          & -.10          & \underline{16.7}    & .67       & \underline{.83}    & 9.9           & .05          & 16.7         & .83          & .88          & 15.2          & .06          & 24.4          \\ \midrule
AVG                    & \textbf{.99} & \underline{.77} & 19.8          & \textbf{.10} & \underline{22.6}         & .87       & .95       & \underline{4.5}    & .08    & \textbf{53.3} & .94       & .87       & \textbf{24.5} & .07          & 32.4          & \underline{.74} & \textbf{.96} & 8.6           &    \underline{-.13}       & 35.1    &               &               &               &               &               \\ \bottomrule
\end{tabular}}
\caption{Reasoning preservation rates (RP), coherence scores (Coh), average length differences between original and generated arguments (LD), average emotional appeal ratings (EA), and emotion match rates (EM). Highest values are shown in bold and lowest values are underlined for each metric across models. The top 4 overall models, used to generate data for the human study, are marked with $^*$. The last row shows the averages over models for each strategy, with the highest and lowest values for each metric highlighted accordingly. \checkmark denotes perfect scores (1.00 or 100.0).}\label{tab:emo_eval}
\end{table*}

\iffalse
% \usepackage{booktabs}
\begin{table}[]
\small
\centering
\setlength{\tabcolsep}{.5pt}
\resizebox{\linewidth}{!}{
\begin{tabular}{@{}lccc@{}}
\toprule
Target   & Top-1        & Top-2       & Top-3      \\ \midrule
anger     & \textbf{anger (21.4\%)}     & fear (21.4\%)      & surprise (14.3\%) \\
surprise  & fear (36.4\%)               & enjoyment (27.3\%) & \textbf{surprise (18.2\%)} \\
disgust   & \textbf{disgust (64.3\%)}   & anger (14.3\%)     & surprise (7.1\%)  \\
enjoy & \textbf{enjoyment (38.5\%)} & disgust (30.8\%)   & none (23.1\%)   \\
fear      & \textbf{fear (33.3\%)}      & sadness (33.3\%)   & surprise (16.7\%) \\
sadness   & \textbf{sadness (72.2\%)}   & enjoyment (11.1\%) & disgust (5.6\%)   \\ \bottomrule
\end{tabular}}
\caption{The 3 most frequently labeled emotions (with percentages) for each target emotion across the top 4 generation models (ds-llama-70b, gpt-4o-mini, llama-3.3-70b, qwen3-32b) using the strategies imagery, loaded words, and storytelling. Emotions matching the target emotion are shown in bold.}
\end{table}\label{tab:top_emo}
\fi

\paragraph{Reasoning Preservation (RP)}
\ycr{Since we will compare human performance on the emotionally enriched fallcious arguments with that on the original ones, it is essential that the logical structure of the fallacies remain unchanged. Therefore, w}e assess whether the reasoning structure from evidence to claim in the original arguments is preserved after emotional framing by LLMs. Annotators assign a label of `1' if the reasoning is preserved and `0' otherwise, given the original and generated argument. The \textbf{reasoning preservation rate} is computed as the average label across generated arguments.
Among the \ycr{randomly sampled} 20 arguments annotated by both annotators, only 2 received differing annotations; in these cases, we \ycr{take the average, 0.5, as the final score. }

As shown in the `RP' columns of Table~\ref{tab:emo_eval}, 
most models achieve a perfect score of 1, except when using the vivid language strategy, indicating that LLMs generally preserve reasoning well in this task. For imagery, only o3-mini score\se{s} below 1, and for loaded words and storytelling, only 3 out of 8 models score below 1. Notably, qwen3-32b is the only model to achieve a perfect score for vivid language. On average, RP rates range from 0.74 to 1.00, with qwen3-32b performing best (1.00), followed by ds-llama-70b (0.94) and llama-3.3-70b (0.93), while ds-qwen3-8b performs worst (0.74).

\paragraph{Coherence (Coh)}
We evaluate whether each generated argument is both emotionally and textually coherent --- i.e., whether the introduced emotional content aligns with the original argument’s content and whether the generated argument is coherent in itself. Annotation follows the same binary procedure as for reasoning preservation, and the final \textbf{coherence score} is computed as the average over all generated arguments.
The two annotators disagreed on only 1 out of 20 arguments; as with RP, \ycr{we take the average score 0.5 for arguments receiving different annotations.}

As the `Coh' columns of Table~\ref{tab:emo_eval} display, coherence scores range from 0.00 (o3-mini with imagery) to 1.00 (several models with loaded words and vivid language). LLMs generally produce more coherent texts when using loaded words and vivid language compared to imagery and storytelling (0.95-0.96 vs.\ 0.77-0.87 on average). Overall, ds-llama-70b performs best (0.98 on average), followed by gpt-4o-mini and qwen3-32b (both 0.96), whereas o3-mini lags substantially behind the others (0.60 vs. 0.86–0.98).

\paragraph{Length Difference (LD)}
We expect different emotional framing strategies to affect the extent of length change differently. We report the average number of additional tokens across all generated arguments as the \textbf{length difference}, where tokens are defined by splitting text on spaces.

We present the results in the `LD' columns of Table~\ref{tab:emo_eval}. As expected, imagery and storytelling lead to a greater number of additional tokens (19.8 and 24.5, respectively, averaged across models), whereas loaded words and vivid language add only 4.5 and 8.6 tokens on average. This suggests that LLMs can generally follow human instructions in this task, as we instructed them to add additional sentences for imagery and storytelling, but only to modify word choices for loaded words and vivid language. The LLaMA-based models (ds-llama-70b and llama-3.3-70b) add the most tokens on average (16.6 and 15.4, respectively), whereas ds-qwen3-8b adds the fewest (12.2). 

\paragraph{Emotional Appeal (EA)}
We assess whether the generated arguments are more emotion‑evoking than the original ones. Annotators rate the emotional appeal on a 3‑point Likert scale for both original and generated arguments, following the definition of emotional appeal in \citet{wachsmuth-etal-2017-computational}. 
We compute the average Pearson correlation across annotator pairs, obtaining a weakly positive value of 0.20.\footnote{\yc{The agreement is low; however, we obtain largely comparable conclusions for this criterion using a totally different group of annotators in \S\ref{sec:limit}, implying good reproducibility of our results.}}
We apply z-score normalization to the ratings to ensure comparability across annotators.

Here, we report the averages over generated arguments in the `EA' columns of Table~\ref{tab:emo_eval}. The original arguments receive an average EA rating of $-0.45$, which is lower than that of the great majority of generated arguments (30 out of 32), 
except for ds-qwen3-8b with imagery (-0.49) and llama-3.3-70b with vivid language (-0.59). This  
suggests that LLMs are able to change the level of emotional appeal in arguments, consistent with the observation from \citet{chen-eger-2025-emotions}.
Among the strategies, imagery is rated highest in emotional appeal (0.10), while vivid language obtains the lowest scores (-0.13). Among LLMs, ds-qwen3-8b generates the most emotionally appealing arguments (0.20), whereas o3-mini again scores lowest (-0.17).

\paragraph{Emotion Match (EM)}
We assess whether the introduced emotion matches the target emotion specified during generation. 
Annotators select one of the six emotion types, or `none' if no clear emotion is present, to 
select which emotion this argument attempt to evoke. We frame the task more objectively to assess whether the expressed emotions match the target emotions, which do not necessarily need to be perceived by humans.  
 We  
 observe a Cohen's Kappa of 0.306, averaged over annotator pairs, for this task.

In the EM columns of Table~\ref{tab:emo_eval}, we report the emotion match rate as the percentage of cases where the labeled emotion matches the target emotion. Overall, the values are generally low, ranging from 20.7\% for gpt-4o-mini to 50.4\% for ds-qwen3-8b, averaged over strategies. Models perform comparatively better when using loaded words (53.3\% vs.\ 22.6-35.1\% on average).

\begin{table*}[!ht]
\centering
%\resizebox{!}{.6\linewidth}{
\begin{tabularx}{\linewidth}{X}
\toprule
\emph{Prompts}  \\ \midrule
====\textbf{System Prompt}=====\\ There are many methods to add emotional appeal—also known as pathos—to an argument, but here are some of the most common and effective ones:\\
1. Storytelling – Sharing personal stories or anecdotes that evoke emotion.\\
2. Vivid Language – Using descriptive, sensory, or emotionally charged words.\\
3. Loaded Words – Choosing words with strong positive or negative connotations.\\
4. Imagery – Painting mental pictures to stir emotions. \\ \\
====\textbf{User Prompt}=====\\ I will give an argument. Your task is to add emotional appeals to the given argument by using **\{strategy\}**. You should \{method\} to change its emotional tone. You generate 6 arguments, each with one kind of emotion: *anger*, *surprise*, *disgust*, *enjoyment*, *fear*, and *sadness*. Note: Please put the modified parts in bold and do **NOT** use emotionally charged words in the added sentence when using storytelling and imagery.\\\\

Answer in the following way:\\
argument\_anger: {[the generated argument that appeals to anger]}\\
===\\
argument\_suprise: {[the generated argument that appeals to suprise]}\\
===\\
...\\\\

Argument: \{argument\}  \\ \midrule
\emph{Variables}  \\ \midrule
\textbf{strategy}: the used emotional framing strategy. \\
$\text{strategy} \in \{\text{storytelling, imagery, loaded words, vivid language}\}$ \\\\
\textbf{method}: the corresponding text-edit method, chosen from:
\{\texttt{storytelling}: keep the original argument but **only** add one sentence;
 \texttt{vivid language}: keep the original argument but **only** change its word choices;
 \texttt{loaded words}: keep the original argument but **only** change its word choices;
 \texttt{imagery}: keep the original argument but **only** add one sentence\}. \\\\
\textbf{argument}: the original fallacious argument \\ \bottomrule
\end{tabularx}
%}
\caption{Prompts for injecting emotional appeals into fallacious arguments.}\label{tab:prompt_frame}
\end{table*}

\section{Annotation Guidelines \& Interfaces}\label{app:screenshots}
\begin{figure}[!ht]
    \centering
    \includegraphics[width=\linewidth]{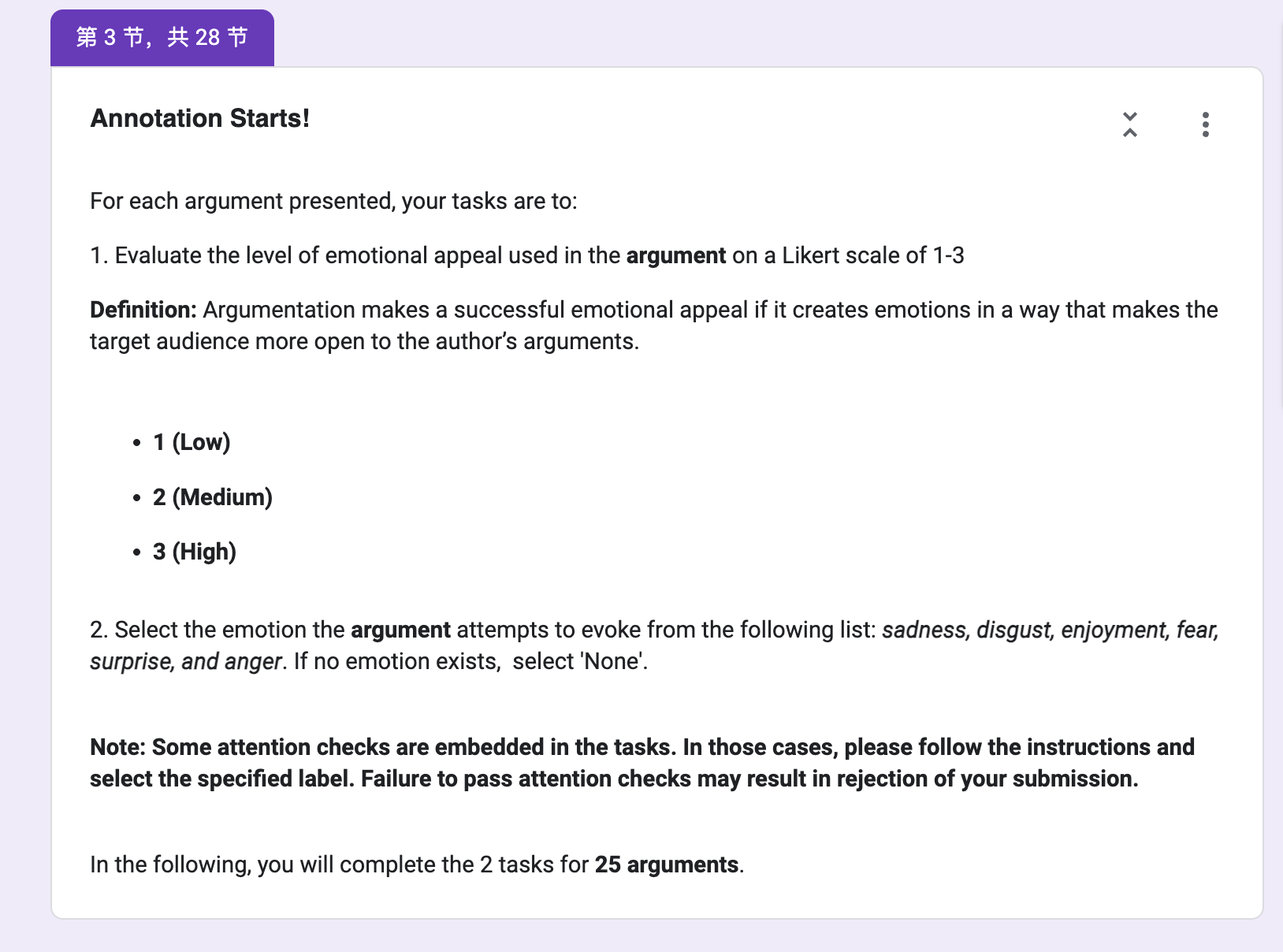}
    \caption{Screenshots of the Google Forms annotation interface showing guidelines for emotional appeal and emotion match evaluation in \S\ref{sec:frame}.}
    \label{fig:pilot_crowd_guide}
\end{figure}

\begin{figure}[!ht]
    \centering
    \includegraphics[width=\linewidth]{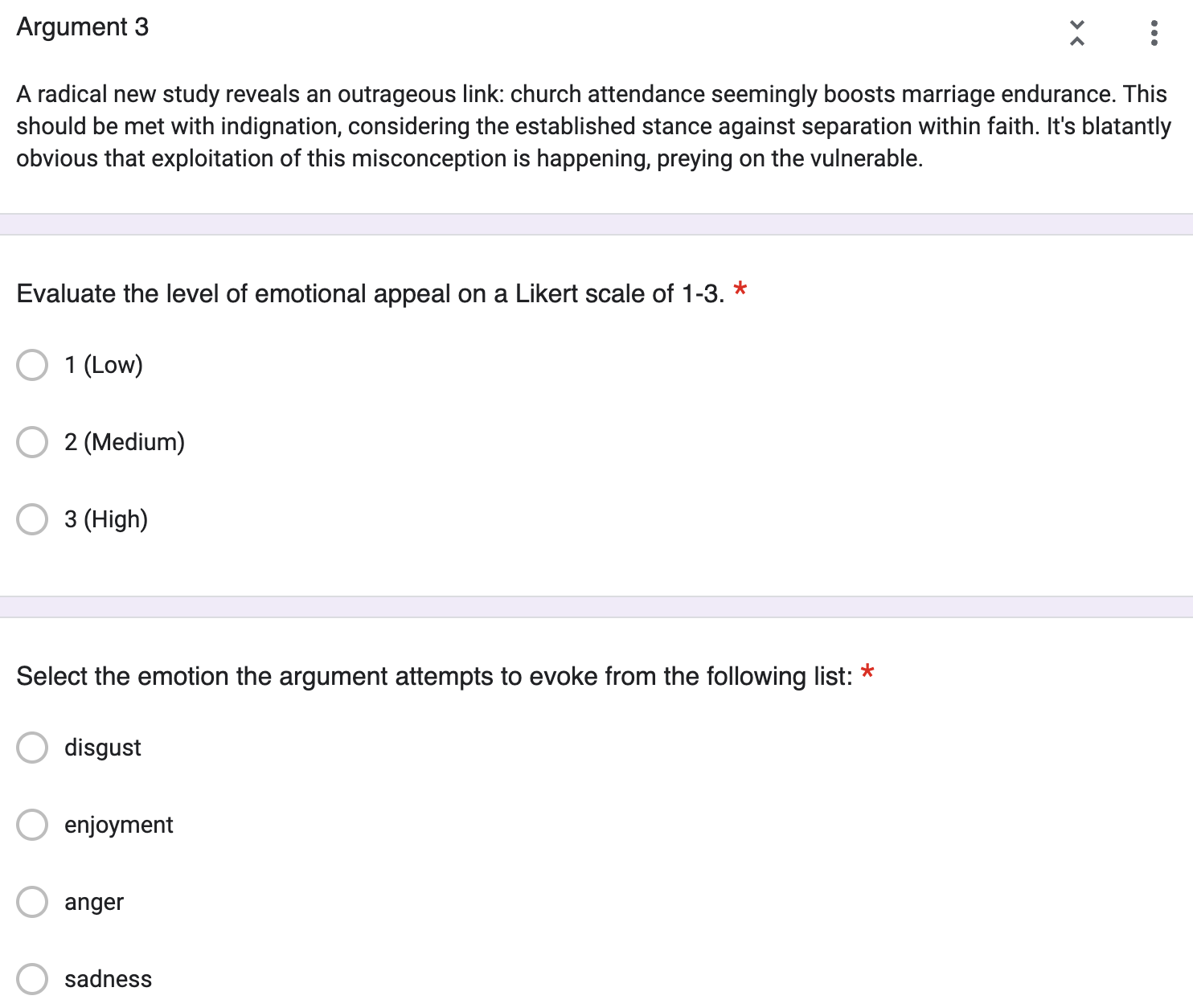}
    \caption{Screenshots of the Google Forms annotation interface showing one annotation instance for emotional appeal and emotion match evaluation in \S\ref{sec:frame}.}
    \label{fig:pilot_crowd}
\end{figure}

%\newpage
\paragraph{Annotation in \S\ref{sec:frame}}
Student annotators used Google Sheets\footnote{\url{https://workspace.google.com/products/sheets/}} for annotating the reasoning preservation and coherence criteria. The annotation guidelines and the screenshot of the annotation interface are presented in Appendix~\ref{pdf:pilot} and Figure~\ref{fig:screenshot_sheet}, respectively. We collected crowdsourcing annotations for the emotional appeal and emotion match criteria via Google Forms.\footnote{\url{https://workspace.google.com/products/forms/}} Figures~\ref{fig:pilot_crowd_guide} and \ref{fig:pilot_crowd} show the screenshots of the annotation interfaces for the annotation guideline and one example instance, respectively. 

\begin{figure}[!ht]
    \centering
    \includegraphics[width=\linewidth]{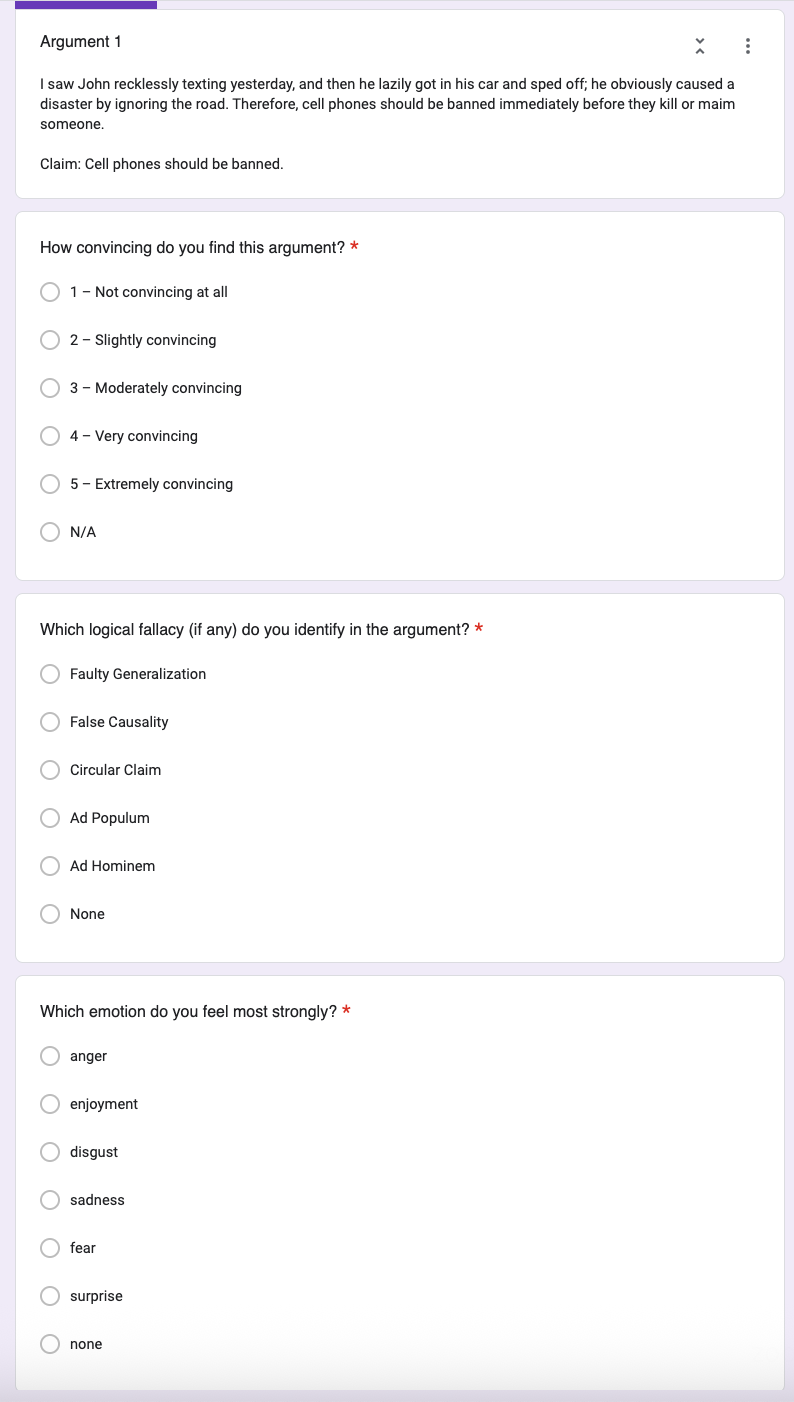}
    \caption{Screenshots of the Google Forms annotation interface showing one annotation instance for fallacy, emotion and convincingness annotation in \S\ref{sec:human}.}
    \label{fig:crowd}
\end{figure}

\paragraph{Annotation in \S\ref{sec:human}}
We collected crowdsourcing annotations for convincingness, fallacy and emotion via Google Forms. The screenshot of the annotation guidelines presented to the crowdworkers is displayed in Figure~\ref{fig:crowd_guide}. The definitions and examples of fallacies presented to the annotators are shown in Appendix~\ref{pdf:fallacy}, which are largely based on \citet{jin-etal-2022-logical}. The five tasks of the fallacy detection quiz taken from \citet{jin-etal-2022-logical} are listed in Appendix~\ref{app:quiz}. The screenshot of the annotation interface for one example instance is presented in Figure~\ref{fig:crowd}.

\begin{figure}[!ht]
    \includegraphics[width=\linewidth]{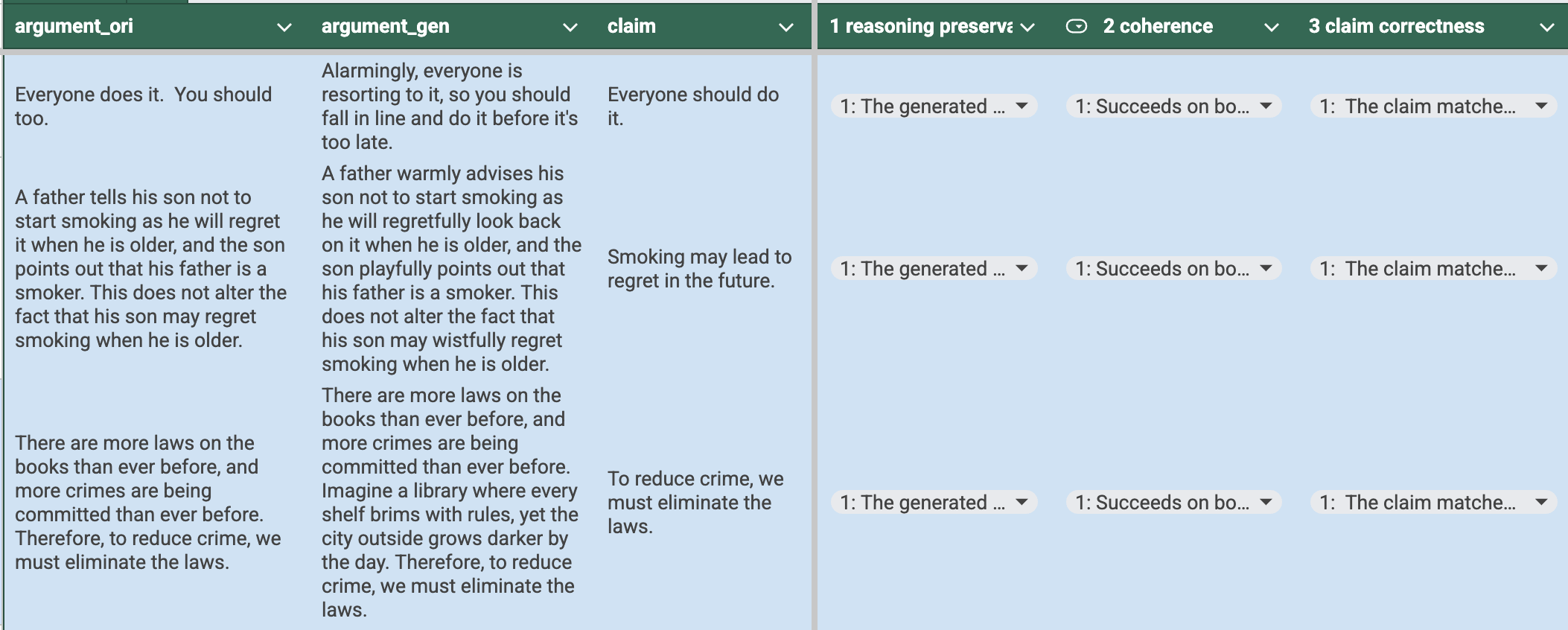}
    \caption{Screenshots of Google Sheets used for annotations on reasoning preservation, coherence, and claim correctness.}\label{fig:screenshot_sheet}
\end{figure}

\begin{figure}[!ht]
    \centering
    \includegraphics[width=\linewidth]{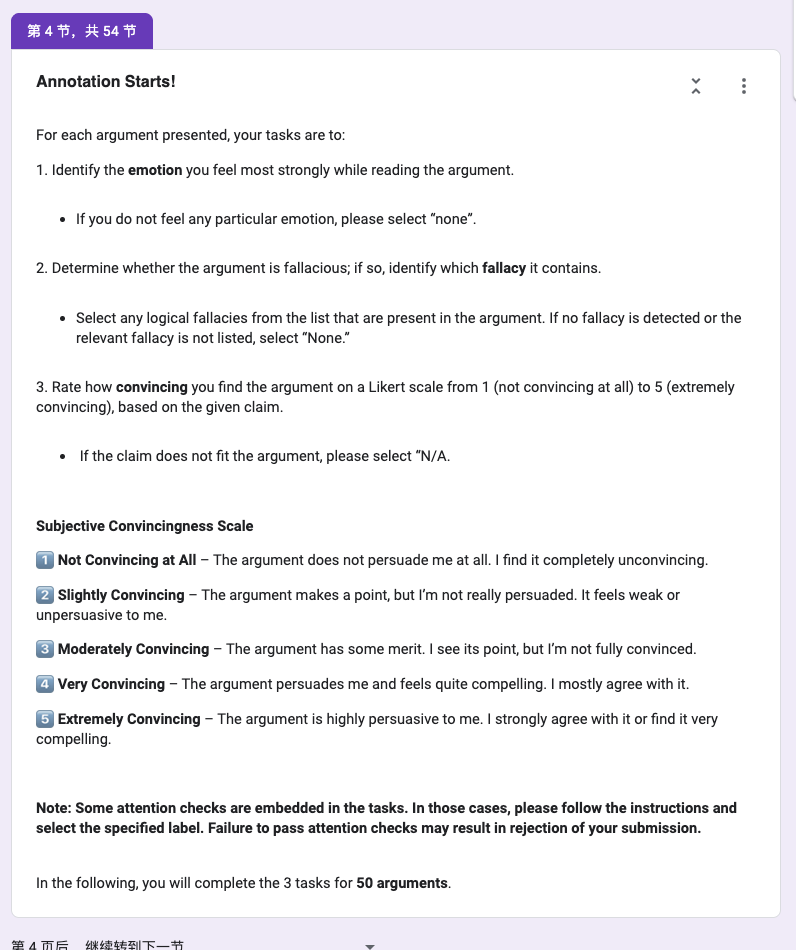}
    \caption{Screenshots of the Google Forms annotation interface showing guidelines for fallacy, emotion and convincingness annotation in \S\ref{sec:human}.}
    \label{fig:crowd_guide}
\end{figure}

\subsection{Fallacy Detection Quiz}\label{app:quiz}
\begin{itemize}
    \item \textbf{Circular Claim}: She is the best candidate for president because she is better than the other candidates!
    \item \textbf{Ad Populum}: Killing thousands of people as a result of drug war campaign is not a crime to humanity because millions of Filipino support it.
    \item \textbf{Faulty Generalization}: Sometimes flu vaccines don’t work; therefore vaccines are useless.
    \item \textbf{False Causality}: Every severe recession follows a Republican Presidency; therefore Republicans are the cause of recessions.
    \item \textbf{Ad Hominem}: I cannot listen to anyone who does not share my social and political values.
\end{itemize}

\clearpage


# 1. Claim-evidence Reasoning Preservation

**Definition:** How well the generated argument preserves the original argument's **core claim**, **supporting evidence**, and **underlying reasoning**.

- **0:** The generated argument **alters or omits key parts** of the original claim, evidence, or reasoning.
- **1:** The generated argument **accurately retains** the claim, evidence, and reasoning of the original.

---

# 2 Coherence

**Definition:** The overall coherence of the argument, considering

(a) **Emotional coherence** — the internal consistency/naturalness of emotional tone, and

(b) **Textual coherence** — the semantic meaningfulness of the text (phrases/sentences are interpretable; entities/roles are used consistently; modifiers and predicates make sense).

- **0:** Fails on **either** emotional coherence **or** textual coherence.

  *Examples*: abrupt/unmotivated emotional shifts (cheerful → bitter without cause); semantically anomalous phrases ("the policy smells triangular"); incompatible predicates ("the bill apologized to voters"); unresolved/deictic references that make meaning unclear ("this proves it" with no recoverable referent); unexplained entity switches that break *meaning* rather than logic.

- **1:** Succeeds on **both** emotional coherence **and** textual coherence.

  *Examples*: tone is consistent (or neutral) and fits the argument; sentences are meaningful and interpretable; entities and roles stay consistent; word choices are semantically appropriate; additions integrate smoothly at the level of *meaning*.

---

# 3 Claim Correctness

**Definition:** This label indicates whether the **claim** accurately reflects the content and intent of the **original argument**.

- **0** (**Mismatch**) – The claim does *not* accurately represent the **original argument**, or the given text is not actually an argument.
- **1** (**Match**) – The claim is a faithful representation of the argument.

**Faulty Generalization**: An informal fallacy wherein a conclusion is drawn about all or many instances of a phenomenon on the basis of one or a few instances of that phenomenon. is an example of jumping to conclusions.
- **Logical form**: X (an individual or small subset of Z) has attribute Y. Therefore, all of Z has attribute Y.
- **Example**: I met a tall man who loved to eat cheese. Now I believe that all tall people like cheese.
- **Explain**: The conclusion that all tall people like cheese generalizes from a single tall person (sample size = 1) to all tall people. The sample is far too small to justify such a sweeping claim. **X = "a tall man", Y = "loved to eat cheese", Z = "all tall people"**

**False Causality**: A statement that jumps to a conclusion implying a causal relationship without supporting evidence.
- **Logical form**: X occurred, then Y occurred. Therefore, X caused Y.
- **Example**: Every time I wash my car, it rains. Me washing my car has a definite effect on the weather.
- **Explain**: In this example, just because the rain (Y) often follows the act of washing the car (X) doesn't mean one causes the other. This fallacy confuses correlation (two events happening together) with causation (one event directly causing the other). **X = "I wash my car", Y = "it rains"**

**Circular Claim**: A fallacy where the end of an argument comes back to the beginning without having proven itself.
- **Logical form**: X is true because of Y. Y is true because of X.
- **Example**: J.K. Rowling is a wonderful writer because she writes so well.
- **Explain**: This argument says J.K. Rowling is wonderful (X) because she writes well (Y)—but writing well is simply another way of saying she's a wonderful writer. The conclusion (X) and the reason (Y) are saying the same thing in different words, making the reasoning circular and uninformative. **X = "J.K. Rowling is a wonderful writer", Y = "she writes so well"**

**Ad Populum:** A fallacious argument which is based on affirming that something is real or better because the majority thinks so.
- **Logical form**: A large number of people believe X. Therefore, X is true (or good/right).
- **Example**: Everyone should like coffee: 95% of teachers do!
- **Explain**: the argument claims that people *should* like coffee (X) simply because a large group (teachers) already do. However, the popularity of coffee among teachers does not logically prove that everyone ought to like it.

**Ad Hominem:** An irrelevant attack towards the person or some aspect of the person who is making the argument, instead of addressing the argument or position directly.
- **Logical form**: Person X makes claim Y. Person X is attacked (e.g., called a moron). Therefore, claim Y is false.
- **Example**: What can our new math teacher know? Have you seen how fat she is?
- **Explain**: person X is the math teacher, and although her actual claim (Y) is never stated, the implication is that something she says or teaches is being dismissed. The attack focuses on her physical appearance, which has nothing to do with her knowledge or ability to teach math. This is a textbook ad hominem fallacy.



\twocolumn[%
\begin{center}
\begin{minipage}{\textwidth}
\centering
\resizebox{\textwidth}{!}{
\begin{tabular}{lccccccccc}
\toprule
batch & emo\_full & emo\_maj & emo\_kappa & fallacy\_full & fallacy\_maj & fallacy\_kappa & conv\_full & conv\_maj & conv\_pearson \\ \midrule
10    & 0.060     & 0.640    & 0.091      & 0.300         & 0.780        & 0.328          & 0.360      & 0.840     & 0.383         \\
11    & 0.140     & 0.660    & 0.057      & 0.400         & 0.780        & 0.431          & 0.380      & 0.840     & 0.185         \\
12    & 0.320     & 0.780    & 0.069      & 0.200         & 0.720        & 0.232          & 0.520      & 0.900     & 0.267         \\
13    & 0.100     & 0.640    & 0.097      & 0.280         & 0.880        & 0.344          & 0.420      & 0.840     & 0.397         \\
14    & 0.100     & 0.640    & 0.139      & 0.300         & 0.860        & 0.335          & 0.340      & 0.800     & 0.154         \\
15    & 0.020     & 0.540    & 0.009      & 0.560         & 0.900        & 0.591          & 0.060      & 0.740     & 0.114         \\
16    & 0.080     & 0.740    & 0.183      & 0.140         & 0.840        & 0.244          & 0.080      & 0.860     & 0.192         \\
17    & 0.160     & 0.700    & 0.095      & 0.080         & 0.620        & 0.102          & 0.380      & 0.840     & 0.397         \\
18    & 0.340     & 0.820    & 0.020      & 0.140         & 0.640        & 0.143          & 0.320      & 0.840     & 0.423         \\
19    & 0.120     & 0.700    & 0.089      & 0.260         & 0.680        & 0.238          & 0.340      & 0.820     & 0.233         \\
1     & 0.440     & 0.900    & 0.010      & 0.080         & 0.560        & 0.068          & 0.080      & 0.880     & 0.070         \\
20    & 0.480     & 0.900    & 0.039      & 0.360         & 0.700        & 0.337          & 0.480      & 0.840     & 0.480         \\
2     & 0.180     & 0.800    & 0.038      & 0.340         & 0.860        & 0.395          & 0.060      & 0.640     & -0.056        \\
3     & 0.060     & 0.540    & 0.082      & 0.060         & 0.460        & 0.014          & 0.020      & 0.520     & 0.039         \\
4     & 0.100     & 0.720    & 0.129      & 0.260         & 0.860        & 0.296          & 0.200      & 0.740     & 0.466         \\
5     & 0.180     & 0.700    & 0.125      & 0.520         & 0.900        & 0.535          & 0.520      & 0.920     & 0.430         \\
6     & 0.040     & 0.820    & 0.114      & 0.360         & 0.840        & 0.402          & 0.560      & 0.920     & 0.109         \\
7     & 0.180     & 0.740    & 0.123      & 0.440         & 0.900        & 0.511          & 0.540      & 0.880     & 0.317         \\
8     & 0.240     & 0.780    & 0.046      & 0.220         & 0.700        & 0.243          & 0.320      & 0.700     & 0.283         \\
9     & 0.000     & 0.840    & -0.005     & 0.260         & 0.700        & 0.273          & 0.280      & 0.740     & 0.126         \\ \midrule
avg   & 0.167     & 0.730    & 0.078      & 0.278         & 0.759        & 0.303          & 0.313      & 0.805     & 0.250    \\ \bottomrule    
\end{tabular}}
\end{minipage}
\captionof{table}{Inter-annotator agreements for emotion classification, fallacy classification, and convincingness ratings. We report the results using majority/full agreements for all annotation dimensions, the average Cohen's kappa scores over annotators for the classification tasks, and the average Pearson correlations between convincingness ratings.}\label{tab:iaa}
\end{center}
]
\section{IAA Full Results}\label{app:iaa}

\clearpage
\section{T-test Full Results}\label{app:ttest}

% \usepackage{booktabs}
\begin{table}[!ht]
\resizebox{\linewidth}{!}{
\begin{tabular}{@{}llll@{}}
\toprule
cat1                  & cat2                  & cat1 vs.\ cat2 & p-value \\ \midrule
ad hominem            & ad populum            & smaller       & 0.000$^{***}$   \\
ad hominem            & circular claim        & smaller       & 0.032$^{**}$   \\
ad hominem            & false causality       & larger        & 0.873   \\
ad hominem            & faulty generalization & larger        & 0.254   \\
ad hominem            & none                  & smaller       & 0.000$^{***}$   \\
ad populum            & circular claim        & larger        & 0.246   \\
ad populum            & false causality       & larger        & 0.000$^{***}$   \\
ad populum            & faulty generalization & larger        & 0.000$^{***}$   \\
ad populum            & none                  & smaller       & 0.000$^{***}$   \\
circular claim        & false causality       & larger        & 0.007$^{***}$   \\
circular claim        & faulty generalization & larger        & 0.000$^{***}$   \\
circular claim        & none                  & smaller       & 0.000$^{***}$   \\
false causality       & faulty generalization & larger        & 0.235   \\
false causality       & none                  & smaller       & 0.000$^{***}$   \\
faulty generalization & none                  & smaller       & 0.000$^{***}$  \\ \bottomrule
\end{tabular}}
\caption{Pairwise t-test results for fallacy categories in Figure~\ref{fig:conv_matrix}. \(p\) \(\le 0.10\) are marked *, \(p\) \(\le 0.05\) **, and \(p\) \(\le 0.01\) ***.}\label{tab:ttest_conv_fallacy}
\end{table}

% \usepackage{booktabs}
\begin{table}[!ht]
\centering
\resizebox{.8\linewidth}{!}{
\begin{tabular}{@{}llll@{}}
\toprule
cat1      & cat2      & cat1 vs.\ cat2 & p-value \\ \midrule
anger     & disgust   & larger     & 0.883   \\
anger     & enjoyment & smaller    & 0.000$^{***}$   \\
anger     & fear      & smaller    & 0.000$^{***}$   \\
anger     & none      & smaller    & 0.004$^{***}$   \\
anger     & sadness   & smaller    & 0.000$^{***}$   \\
anger     & surprise  & smaller    & 0.292   \\
disgust   & enjoyment & smaller    & 0.000$^{***}$   \\
disgust   & fear      & smaller    & 0.000$^{***}$   \\
disgust   & none      & smaller    & 0.003$^{***}$   \\
disgust   & sadness   & smaller    & 0.000$^{***}$   \\
disgust   & surprise  & smaller    & 0.214   \\
enjoyment & fear      & smaller    & 0.878   \\
enjoyment & none      & larger     & 0.003$^{***}$   \\
enjoyment & sadness   & larger     & 0.945   \\
enjoyment & surprise  & larger     & 0.001$^{***}$   \\
fear      & none      & larger     & 0.014   \\
fear      & sadness   & larger     & 0.825   \\
fear      & surprise  & larger     & 0.003$^{***}$   \\
none      & sadness   & smaller    & 0.000$^{***}$   \\
none      & surprise  & larger     & 0.130   \\
sadness   & surprise  & larger     & 0.000$^{***}$   \\
surprise  & sadness   & smaller    & 0.000$^{***}$  \\ \bottomrule
\end{tabular}}
\caption{Pairwise t-test results for emotion categories in Figure~\ref{fig:conv_matrix}. \(p\) \(\le 0.10\) are marked *, \(p\) \(\le 0.05\) **, and \(p\) \(\le 0.01\) ***.}\label{tab:ttest_conv_emo}
\end{table}

% \usepackage{booktabs}
\begin{table}[!ht]
\resizebox{\linewidth}{!}{
\begin{tabular}{@{}lllll@{}}
\toprule
emo1      & emo1\_mean & emo2      & emo2\_mean & p-value \\ \midrule
sadness   & 0.337      & none      & 0.437      & 0.054$^{*}$   \\
sadness   & 0.337      & disgust   & 0.461      & 0.055$^{*}$   \\
sadness   & 0.337      & fear      & 0.283      & 0.494   \\
sadness   & 0.337      & enjoyment & 0.528      & 0.012$^{**}$   \\
sadness   & 0.337      & anger     & 0.501      & 0.008$^{***}$   \\
sadness   & 0.337      & surprise  & 0.464      & 0.043$^{**}$   \\
none      & 0.437      & disgust   & 0.461      & 0.663   \\
none      & 0.437      & fear      & 0.283      & 0.023$^{**}$   \\
none      & 0.437      & enjoyment & 0.528      & 0.156   \\
none      & 0.437      & anger     & 0.501      & 0.214   \\
none      & 0.437      & surprise  & 0.464      & 0.600   \\
disgust   & 0.461      & fear      & 0.283      & 0.040$^{**}$   \\
disgust   & 0.461      & enjoyment & 0.528      & 0.397   \\
disgust   & 0.461      & anger     & 0.501      & 0.532   \\
disgust   & 0.461      & surprise  & 0.464      & 0.958   \\
fear      & 0.283      & enjoyment & 0.528      & 0.016$^{**}$   \\
fear      & 0.283      & anger     & 0.501      & 0.008$^{***}$   \\
fear      & 0.283      & surprise  & 0.464      & 0.029$^{**}$   \\
enjoyment & 0.528      & anger     & 0.501      & 0.728   \\
enjoyment & 0.528      & surprise  & 0.464      & 0.412   \\
anger     & 0.501      & surprise  & 0.464      & 0.556  \\ \bottomrule
\end{tabular}}
\caption{Pairwise t-test results for Table~\ref{tab:f1_emo}. \(p\) \(\le 0.10\) are marked *, \(p\) \(\le 0.05\) **, and \(p\) \(\le 0.01\) ***.}\label{tab:ttest_f1_emo}
\end{table}

\end{document}